\documentclass[final,3p,times]{elsarticle}
\usepackage{graphicx} 
\usepackage[utf8]{inputenc}
\usepackage{caption}
\usepackage{subcaption}
\usepackage{siunitx}
\usepackage{float} 
\usepackage{setspace}
\usepackage{amsmath}
\usepackage{cite}
\usepackage[labelfont=bf,labelsep=colon]{caption}

\def\tsc#1{\csdef{#1}{\textsc{\lowercase{#1}}\xspace}}
\tsc{WGM}
\tsc{QE}
\tsc{EP}
\tsc{PMS}
\tsc{BEC}
\tsc{DE}

\makeatletter
\def\ps@pprintTitle{%
  \def\@oddfoot{}%
  \def\@evenfoot{}%
}
\makeatother

\begin{document}
\let\WriteBookmarks\relax
\def\floatpagepagefraction{1}
\def\textpagefraction{.001}

\title{FGM optimization in complex domains using Gaussian process regression based profile generation algorithm }

\author{Chaitanya Kumar Konda\textsuperscript{a}} 
\author{Piyush Agrawal\textsuperscript{a}}
\author{Shivansh Srivastava\textsuperscript{a}}
\author{Manish Agrawal\textsuperscript{a}\textsuperscript{,*}}
\ead{manish.agrawal@iitrpr.ac.in}

\cortext[cor1]{Corresponding author.}

\address[1]{Department of Mechanical Engineering, Indian Institute of Technology Ropar, Rupnagar-140001, Punjab, India}

\begin{abstract}

This manuscript addresses the challenge of designing functionally graded materials (FGMs) for arbitrary-shaped domains. Towards this goal, the present work proposes a generic volume fraction profile generation algorithm based on Gaussian Process Regression (GPR). The proposed algorithm can handle complex-shaped domains and generate smooth FGM profiles while adhering to the specified volume fraction values at boundaries/part of boundaries. The resulting design space from GPR comprises diverse profiles, enhancing the potential for discovering optimal configurations. Further, the algorithm allows the user to control the smoothness of the underlying profiles and the size of the design space through a length scale parameter. Further, the proposed profile generation scheme is coupled with the genetic algorithm to find the optimum FGM profiles for a given application. To make the genetic algorithm consistent with the GPR profile generation scheme, the standard simulated binary crossover operator in the genetic algorithm has been modified with a projection operator. We present numerous thermoelastic optimization examples to demonstrate the efficacy of the proposed profile generation algorithm and optimization framework.

\end{abstract}


\maketitle

\section{Introduction}

Functionally graded materials (FGMs) represent a class of advanced composite materials characterized by a gradual variation in the composition of their constituents throughout the material domain. Unlike conventional composite materials, which undergo failure due to delamination between the matrix and fibers—particularly under high-temperature conditions \citep{gong2022temperature} caused by mismatched thermal expansion coefficients—FGMs effectively mitigate such issues through their continuous compositional gradation. This unique property of FGM not only enhances structural integrity but also improves the overall thermal and mechanical performance of the material. As a result, FGMs are employed in high-performance sectors such as aerospace, biomedical engineering, and defense technologies \citep{boggarapu2021state}. FGMs offer numerous advantages such as high fracture toughness, reduction in the plane and through the thickness transverse stresses, enhanced performances in thermal barriers, improved residual stress distribution, etc \citep{chi2006mechanical,bhavar2017review,koizumi1995overview}.

The advantages of FGMs stem from the optimal tailoring of constituent material gradation.  Hence, finding the suitable volume fraction distribution of an FGM is a fundamental step in designing FGMs for any specific applications. In the existing literature, there are numerous works on optimizing the volume fraction distribution of FGMs for various objectives. On this note, \citep{cho2002optimal,goupee2006two,nemat2009reduction,ding2018optimization,correia2019multiobjective} worked on minimizing the stress or mass of the FGM plates due to thermal and mechanical loading by tailoring the volume fraction composition of the FGM plate. Ashjari et al. \citep{ashjari2017multi} designed an FGM sandwich panel to minimize mass and deflection under mechanical loading with stress constraints. Other work, \citep{alshabatat2014optimization,roque2016differential} demonstrates the role of volume fraction distribution in FGM beams under the objective minimization of free vibration frequency.

A prerequisite for optimizing FGMs is the creation of a well-defined design space composed of diverse volume fraction profiles. In the existing literature, numerous profile generation algorithms have been developed to construct FGM profiles and aid in identifying optimal designs. Among the most commonly used are the power law, sigmoidal law, exponential law, and trigonometric law. For instance, Moita et al. \citep{moita2018material} employed the power law to design FGM plate-shell structures, optimizing objectives such as mass, displacement, fundamental frequency, and critical load. Power law-based formulations have also been used in multi-objective optimization problems. For example, Correria et al. \citep{correia2019multiobjective} optimized FGM plates by minimizing mass, material cost, and failure stress under thermo-mechanical loading. Due to its mathematical simplicity and ease of implementation, the power law remains a popular choice in many studies involving FGM structures \citep{chi2006mechanical,birman2007modeling,zhang2012designing,kamarian2014application}. Other variations, such as the exponential law \citep{reddy2014three}, sigmoidal law \citep{chi2006mechanical,ding2018optimization}, and three-parameter power law \citep{yas2014application} offer alternative gradation profiles. Beyond these classical approaches, several researchers have introduced more flexible and adaptable profile generation algorithms. Alshabatat et al. \citep{alshabatat2014optimization} proposed two novel gradation laws—a four-parameter power law and a five-parameter trigonometric law—that offer enhanced control over material distribution through additional tuning parameters. In a similar vein,  Hussein et al. \citep{hussein2017multi} utilized a polynomial expansion based on spatial coordinates to generate smooth and continuous FGM profiles, enabling a coordinate-dependent formulation of material gradation. In addition to these mathematical expressions, grid-based approaches have also been explored in the literature \citep{cho2002optimal,goupee2006two}. For example, Goupee et al. \citep{goupee2006two} discretized the material domain into a finite set of grid points, assigning volume fractions at each point. To ensure smooth transitions in the gradation of constituents, partial derivatives were specified at grid points to maintain  $C^o$ and $C^1$ continuity across the domain. Moreover, Taheri et al. \citep{taheri2014simultaneous} applied Non-Uniform Rational B-Splines (NURBS) to simultaneously model the geometry and material gradation of FGM structures, aiming to optimize the eigen frequencies. Although these existing FGM profile generation algorithms are simple and computationally efficient, they are predominantly tailored only for the standard geometry problem, such as squares and rectangles.

Genetic Algorithm (GA) is one of the widely employed optimization algorithms to determine the optimal design of FGMs. It belongs to the class of gradient-free or random search methods. The robustness of GA makes it applicable to many kinds of problems. Authors \citep{goupee2006two,chiba2012optimisation,ding2018optimization} deployed GA to obtain the optimal FGM profile of a FGM plate under thermo-mechanical loading to mitigate the thermal stresses. Nguyen et al. \citep{nguyen2017optimal} used GA to optimally design thin-walled FGM beams to withstand lateral and flexural-torsional buckling loads. Non-dominated sorting genetic algorithm (NSGA-II) is used by Ashjari et al. \citep{ashjari2017multi} to perform multi-objective optimization of minimizing mass and deflection of FGM sandwich panels under mechanical loading. In a typical GA based optimization framework, the evaluation of the design is guided by a fitness score, which serves as the basis for ranking and selection. In this work, fitness scores are computed using finite element simulations of FGMs. The buckling load in a thin I-section FGM beam is evaluated \citep{nguyen2017optimal} utilizing the FEM. Bhangale et al.\citep{bhangale2006thermoelastic} used the FEM to study the vibration and buckling behavior of a FGM sandwich beam having a constrained viscoelastic layer in a thermal environment. In the literature, various attempts have been made using gradient-based methods for FGM optimization \citep{hussein2017multi, taheri2014simultaneous, liu2003optimization, chen2005thermomechanically}.

As stated above, in the literature, most of the existing work on the FGM optimization is restricted to simple geometries. To overcome the limitations of the existing work, here we propose a profile generation algorithm based on the concept of Gaussian process regression (GPR). The basis of the GPR is the conditional multivariate Gaussian distribution obtained from the Bayesian framework. The prior distribution in the Bayesian framework is taken as a multivariate Gaussian distribution, with the covariance matrix being calculated from a kernel function. Readers can refer Wang et al. \citep{wang2023intuitive} for details on the GPR. GPR has been used in many engineering applications, with the most prominent one being to perform regression tasks in the field of machine learning \citep{williams2006gaussian}. The regression modeling using GPR has been extensively used in surrogate modeling across various domains. For instance, reliability analysis of complex structures excluding FEM \citep{su2017gaussian}, predictions of stress intensity factor to assess crack propagation \citep{loghin2019augmenting}, and the computationally expensive optimization problems \citep{luo2023dynamic} deploy the GPR as a surrogate model. Apart from surrogate modeling, Choubey et al. \citep{choubey2024novel} used GPR to generate the temperature field in the various domains. 

 In this work, our objective is to find the optimal material distribution in complex FGM domains. Towards this objective, we employ GPR  as the profile generation scheme along with the genetic algorithm framework to find the optimal material distribution under thermo-mechanical loading conditions. In the GPR-based algorithm, first, the conditional multivariate Gaussian is obtained for the specified volume fraction at the boundary/part of the boundary. The random profiles, henceforth, are generated from this conditional multivariate Gaussian distribution. The proposed profile generation algorithm generates smooth profiles for arbitrary-shaped domains, and naturally incorporates volume fraction constraints (pure ceramic or pure metallic region) within the boundaries of the FGM. These random profiles generated from the algorithm are utilized to form the underlying design space for optimization purposes. The size of the design space and the smoothness of the profiles can be controlled by the length hyperparameter.

As stated above, we adopt the use of the genetic algorithm in conjunction with the proposed profile generation scheme for optimization. In the literature, to the best of our knowledge, this is the first such attempt to combine the GPR with the GA framework. The deployment of the standard GA is not feasible in our setting, since the crossover operator causes a disruption in the smoothness of profiles. This disruption in the smoothness is rectified by applying a projection operator after crossover. To ensure consistency between the mutation operation and the profile generation framework, a Gaussian mutation consistent with the GPR framework is incorporated. Thus, proposed modifications to the crossover and mutation operators ensure that profiles are consistent with the design space at each phase of GA. The efficacy of the proposed optimization approach combining GA and GPR is demonstrated through several numerical examples involving complex geometries and various design constraints.

The remainder of this manuscript is organized as follows: Section~\ref{PGA} outlines the steps of the profile generation algorithm used to construct the FGM design space. The thermoelastic FEA methodology for FGM is presented in Section~\ref{FEM}. Section~\ref{sec_GA} then describes the GA-based optimization framework and its integration with the profile generation algorithm. Finally, Section~\ref{sec_NE} provides numerical examples that demonstrate the effectiveness of the proposed approach.

\section{Profile Generation Algorithm} \label{PGA}
In this section, we present the FGM profile generation algorithm as a first step towards finding the tailor-made profile for a specific application. This profile generation algorithm will enable the formulation of the underlying space, which will be used in conjunction with the optimization framework in the later section. In the profile generation algorithm, we strive to achieve three key characteristics: 1) The profile generation algorithm should be applicable to any arbitrary-shaped domain. This is a crucial requirement for the generality of the generation scheme. 2) There should be significant variance in the design profiles, i.e., the underlying design space should be sufficiently large. The larger design space is desirable to maximize the performance of the optimum profile. 3) The profile generated should be continuous and smooth in nature, and the designer should be able to control the smoothness based on the underlying application. The presence of unnecessary non-smooth designs leads to increased complexity of optimization. 

\begin{figure}
    \centering
    \includegraphics[width=0.55\textwidth]{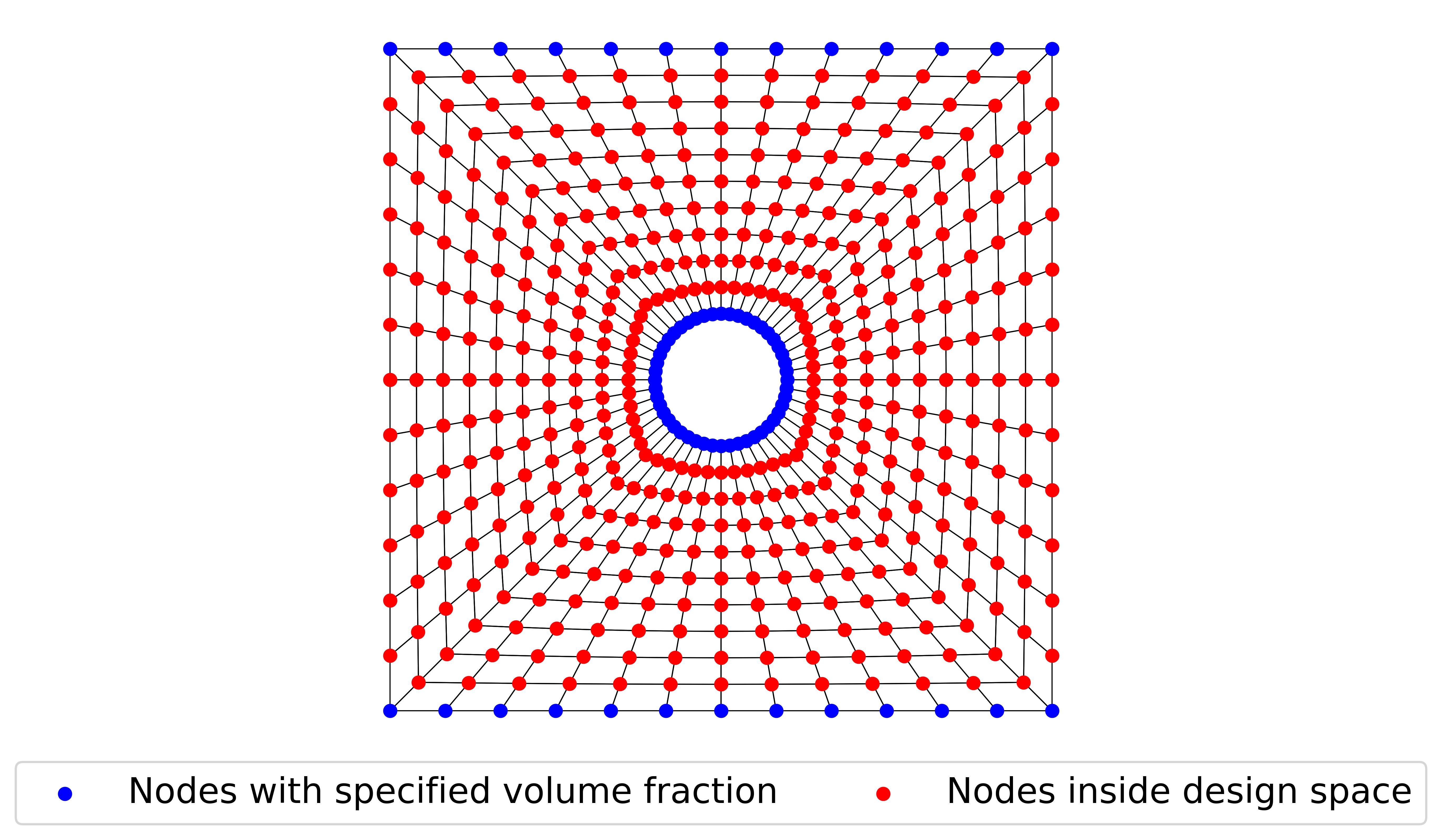} 
    \caption{Sample domain discretized into the elements and nodes. The volume fraction at  nodes are generated from the multivariate Gaussian distribution.}
    \label{GPR_nodes}
\end{figure}

To achieve the stated three objectives, we utilize Gaussian Process regression (GPR) for the generation of the FGM profiles. We consider that the volume fraction of one constituent of FGM is a Gaussian random function of spatial variables (\(x\) and \(y\)). Once the distribution of one constituent is known, the volume fraction of the other constituent can be calculated trivially. As a first step in the algorithm, we discretize the domain into a finite number of elements/nodes. For example, the discretization of a square plate with a circular hole domain is shown in Fig. \ref{GPR_nodes}. We generate the values of the volume fraction at the nodal points of the domain through a multivariate Gaussian distribution.

 The volume fraction in the nodes can be represented : 
\begin{equation}
\begin{aligned}
    V(X) \sim \mathcal{N}(\mu(\boldsymbol{X}),\boldsymbol{K}),
     \label{eq:prior_dist}
\end{aligned}
\end{equation}
\noindent where $V(X)$ represents the volume fraction distribution function of spatial coordinates in the domain, the multivariate Gaussian distribution is given by $\mathcal{N}$, $\mu(X)$ is the mean of the volume fraction distribution, choosen to be zero, and $\boldsymbol{K}$ is the covariance matrix.

Further, the linear interpolation is used to define the volume fraction at a point within any element. The covariance matrix can be defined with the help of the kernel functions. In literature, various choices of the kernel functions are presented, with the radial basis function (RBF) being the most popular one. The RBF is defined by the two hyperparameters \(l\) and \(\sigma\), given by the following expression:

\begin{equation}\label{RBF}
\begin{aligned}
    K_{ij} = k({X}_i, {X}_j) = \sigma^2 \exp\left( -\frac{\|{X}_i - {X}_j\|^2}{2l^2} \right),
\end{aligned}
\end{equation}
 where $\sigma$ is the vertical parameter that determines the spread of the values, thereby determining the magnitude of the values generated. While $l$ is a characteristic length scale that determines the smoothness of the gradation in an FGM profile, smoothness increases with an increase in the length scale value. Further, the FGM profiles obtained from the lower length scales have a higher amount of variability compared to the larger length scales. The design space with the large length scale is a subset of the design space with a smaller length scale. However, decreases in the value of the length parameter result in an increase in the optimization complexity.   So, a proper choice of length scale has to be made based on the dimensions of the geometry, design constraints, and gradation requirements. As a result, in the numerical examples shown in the section \ref{sec_NE}, we have used different $l$ along with different $\sigma$ values based on the problem requirements.

 In most of the FGM problems, the volume fraction at certain boundaries is predefined as shown in Fig. \ref{GPR_nodes}. For example, in the case of a high-temperature application, the boundary with a high temperature value consists of pure ceramic. Thus, it is required to incorporate these predefined constraints on the design space. To do so, in line with the standard GPR technique, we use a Bayesian framework to obtain the conditional Gaussian distribution \( p(\boldsymbol{f} \mid \boldsymbol{f_{b}}, \boldsymbol{X}, \boldsymbol{X}_b) \). Here, \(\boldsymbol{X}\) is a vector containing the coordinates of all nodes and \(\boldsymbol{f}\) denotes the volume fraction value,  \(\boldsymbol{X_{b}}\) contains the coordinates of boundary nodes where the volume fraction is prescribed. While \(\boldsymbol{f_{b}}\) denotes these prescribed volume fraction values.  
 

The final posterior distribution is obtained by the following equations:

\begin{equation}
\begin{aligned}
    \boldsymbol{f \mid f_{b}}, \boldsymbol{X}, \boldsymbol{X}_b &\sim \mathcal{N} \left( \mu^{\star}(\boldsymbol{X}),\boldsymbol{K^{\star}} \right),
\end{aligned}
\label{eq:f_star_given_f}
\end{equation}
where,
\begin{equation}
\begin{aligned}
    \boldsymbol{\mu^{\star}} &= \mu(\boldsymbol{X})+\boldsymbol{K}_*^T (\boldsymbol{K}+ \sigma^{2}_{n}\boldsymbol{I})^{-1}( \boldsymbol{f_{b} - \mu(X_{b})}), \\
    \boldsymbol{K^{\star}} &= \boldsymbol{K}_{**} - \boldsymbol{K}_*^T (\boldsymbol{K_{b}}+ \sigma^{2}_{n}\boldsymbol{I})^{-1} \boldsymbol{K}_*,
\end{aligned}
\label{K_exp}
\end{equation}
where 
$\boldsymbol{K}$ denotes the covariance matrix among the boundary nodes, $ \boldsymbol{K}_*$ denotes the covariance matrix between the boundary nodes and the prediction nodes, and $\boldsymbol{K}_{**}$ denotes the covariance matrix among the prediction nodes. $\boldsymbol{\mu^{\star}}$ denotes the mean of the posterior distribution with variance $\boldsymbol{K^{\star}}$. The term $\sigma^{2}_{n}$ is required to ensure the non-singularity of the covariance matrix and represents the assumed noise variance. $\boldsymbol{I}$ is the identity matrix. Also, note that the value of $\sigma^{2}_{n}$ need to chosen a small number (in our case $\sigma^{2}_{n}$ = $10^{-3}$) to satisfy the boundary volume fraction constraints accurately.

\begin{figure*}[ht]
    \centering
    
    \begin{subfigure}{\textwidth}
        \centering
        \begin{subfigure}{0.9\textwidth}
            \includegraphics[width=\linewidth]{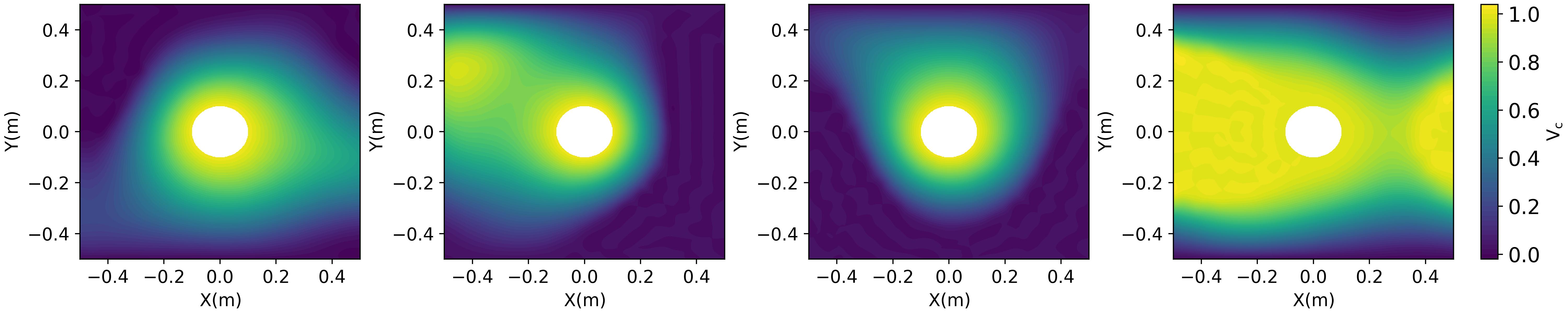}
        \end{subfigure}
        \caption{}
    \end{subfigure}
    
    \vspace{0.5em} 
    
    \begin{subfigure}{\textwidth}
        \centering
        \begin{subfigure}{0.9\textwidth}
            \includegraphics[width=\linewidth]{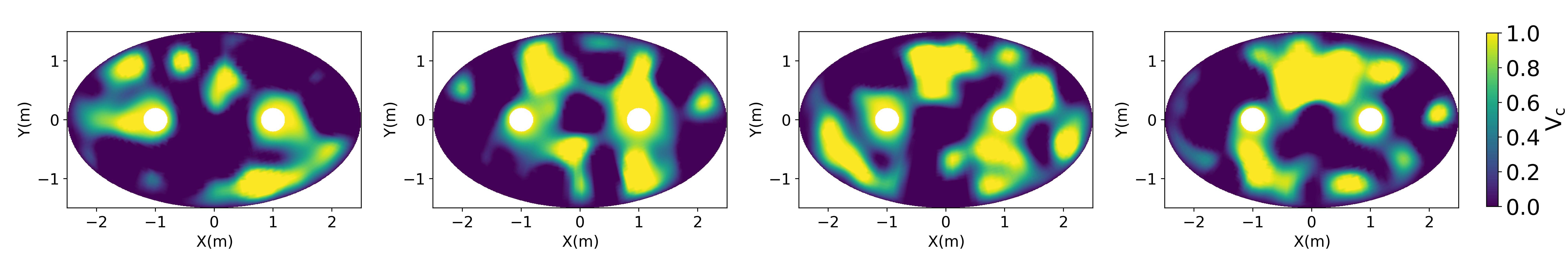}
        \end{subfigure}
        \caption{}
    \end{subfigure}

    \vspace{0.5em} 
    
    \begin{subfigure}{\textwidth}
        \centering
        \begin{subfigure}{0.9\textwidth}
            \includegraphics[width=\linewidth]{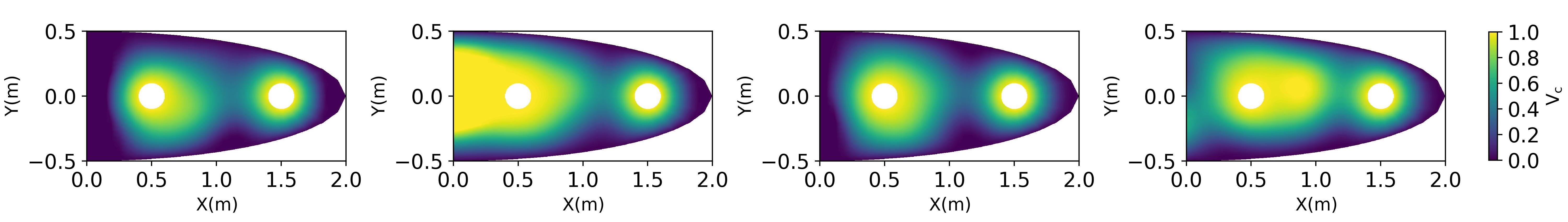}
        \end{subfigure}
        \caption{}
    \end{subfigure}
    
    \caption{Posterior samples of FGM profiles obtained using GPR, satisfying pure metal $(V_{c} = 0)$ and pure ceramic $(V_{c} = 1)$ boundary conditions: (a) square plate with circular hole, having pure metal at the bottom and top edge, and pure ceramic at the periphery of hole, (b) ellipse geometry with two circular holes, having pure metal at the periphery of ellipse and pure ceramic at the periphery of holes, and (c) half ellipse geometry with two circular holes, having pure metal at the outer curved surface and pure ceramic at the periphery of holes. }
    \label{posterior}
\end{figure*}

From the above posterior distribution, the obtained FGM profiles are smooth in nature and satisfy the prescribed boundary values. Fig. \ref{posterior} shows a few examples of the posterior profiles obtained by enforcing specific volume fraction values at the boundary nodes for three different domains.

\section{Finite Element Analysis of Thermoelastic problems} \label{FEM}
This section presents the governing equations for thermoelastic problems, followed by a variational formulation of the governing equations. In this analysis, we assume that metal and ceramic are linearly elastic material and their material properties are independent of temperature. Also, the domain under consideration is initially in a stress-free state, and the metal-ceramic interfaces are perfectly bonded.

\subsection{Governing equations}
The steady-state temperature distribution within the domain ($\Omega$) is governed by the thermal equilibrium equation in conjunction with Fourier’s law of heat conduction.
\begin{subequations} 
    \begin{align}
        -\nabla \cdot \boldsymbol{q} + Q &= 0 \quad \text{on } \Omega, \label{eq:thermal} \\
        \boldsymbol{q} &= -\kappa \nabla \theta, \label{eq:fourier}
    \end{align}
\end{subequations}

\noindent where $\boldsymbol{q}$ denotes the heat flux vector, $Q$ is the internal heat generation per unit volume, $\kappa$ is the thermal conductivity, and $\theta$ represents the temperature rise from the reference state.\\

The following thermal boundary conditions are imposed as: heat flux ($\hat{q}$) specified in the direction normal to boundary $\Gamma_q$, convective heat transfer specified on the boundary $\Gamma_c$, and the temperature $\hat{\theta}$ is set on boundary $\Gamma_\theta$\begin{subequations}
   \begin{align}
    -\boldsymbol{q} \cdot \boldsymbol{n} + h (\theta - \theta_{\infty}) &= 0 \quad \text{on } \Gamma_c, \label{TBC1} \\
        -\boldsymbol{q} \cdot \boldsymbol{n} &= \hat{q} \quad \text{on } \Gamma_q, \\
            \theta &= \hat{\theta} \quad \text{on } \Gamma_{\theta}, 
    \end{align}
\end{subequations}

\noindent where, $\boldsymbol{n}$ is the unit vector normal to the boundary, $h$ is the convective heat transfer coefficient of the surroundings and $\theta_{\infty}$ is the ambient temperature.\\

The thermoelastic deformation of a body is governed by the following equilibrium and constitutive relations, along with the temperature field obtained from the thermal analysis
\begin{subequations}
    \begin{align}
        \nabla \cdot \boldsymbol{\sigma} + \boldsymbol{b} &= 0 \quad \text{on } \Omega, \label{elastic} \\
        \boldsymbol{\sigma} &= \boldsymbol{\sigma_m} - \beta \theta \mathbf{I}, \\
        \boldsymbol{\sigma_m} &= \lambda (\operatorname{tr} \boldsymbol{\epsilon}) \mathbf{I} + 2\mu \boldsymbol{\epsilon},
    \end{align}
\end{subequations}

\noindent where, $\boldsymbol{\sigma}$ is the  Cauchy stress tensor, $\boldsymbol{\sigma_m}$ is the isothermal stress tensor, $\boldsymbol{b}$ is the body force, $\beta$ is the stress-temperature coefficient, $\lambda$ and $\mu$ are the Lame constants, $\boldsymbol{I}$ is the identity tensor. The infinitesimal strain tensor $\boldsymbol{\epsilon}$ in terms of displacement vector $\boldsymbol{u}$ is given by:
\begin{equation}
   \boldsymbol{\epsilon} = \frac{1}{2} \left( (\nabla \boldsymbol{u}) + (\nabla \boldsymbol{u})^T \right),
\end{equation}

\noindent and the thermal stress coefficient $\beta$ is defined as:

\begin{center}
\begin{tabular}{@{}c@{\hspace{3cm}}c@{}}
Plane Stress:
$\displaystyle \beta = \frac{E \alpha}{1 - \nu}$, 
&
Plane Strain: 
$\displaystyle \beta = \frac{E \alpha}{1 - 2\nu}$, \\
\end{tabular}
\end{center}
where $\alpha$ is the coefficient of thermal expansion and $\nu$ is Poisson's ratio. The mechanical boundary conditions are imposed in the form of specified traction $\bar{t}$ on the boundary $\Gamma_t$ and set displacement $\boldsymbol{u}$ on the boundary $\Gamma_u$: \begin{subequations}
    \begin{align}
    \boldsymbol{u} &= \boldsymbol{u}_0 \quad \text{on } \Gamma_u, \\
        \boldsymbol{t} &= \boldsymbol{\bar{t}} \quad \text{on } \Gamma_t. 
    \end{align}
\end{subequations}

\subsection{Variational formulation}

In this section, we present the weak form obtained by the governing equations. The weak form of the heat conduction equation to perform thermal analysis is obtained by multiplying the variation $\theta_\delta$ to Eq.~\eqref{eq:thermal} and further integrating by parts to get:
\begin{equation} \label{Therm_Var1}
    \int_{\Omega} \nabla \theta_{\delta} \cdot (k \nabla \theta) \, d\Omega
    - \int_{\Omega} \theta_{\delta} Q \, d\Omega
    - \int_{\Gamma_q} \theta_{\delta} \hat{q} \, d\Gamma
    + \int_{\Gamma_c} \theta_{\delta} h (\theta - \theta_{\infty}) \, d\Gamma = 0,
    \quad \forall \theta_{\delta}, 
\end{equation}

Similarly, the weak form of elastic equilibrium equations to perform elastic analysis is obtained by multiplying the variation $\boldsymbol{u_\delta}$ to Eq.~\eqref{elastic} and further integrating by parts to get:
\begin{equation} \label{FE_Elastic}
\int_{\Omega} \boldsymbol{\sigma}_m : \nabla \boldsymbol{u}_\delta \, d\Omega 
- \int_{\Omega} \boldsymbol{u}_\delta \cdot \boldsymbol{b} \, d\Omega 
+ \int_{\Gamma_t} \boldsymbol{u}_\delta \cdot \boldsymbol{t} \, d\Gamma 
+ \int_{\Omega} \beta \theta \nabla \cdot \boldsymbol{u}_\delta \, d\Omega = 0, 
\quad \forall \boldsymbol{u}_\delta.
\end{equation}

The material properties at the Gauss points are calculated using the rule of mixture given by:
\begin{equation}
P(x, y) = P_m (1-V_c(x, y)) + P_cV_c(x, y)),
\end{equation}

\noindent where, \( P(x, y) = \) Property at the Gauss point, 
\( P_m = \) Property of the metallic phase, 
\( P_c = \) Property of the ceramic phase,  
\( V_c = \) Volume fraction of ceramic phase at the Gauss point.

For nine-noded quadrilateral elements  3x3 points Gauss quadrature rule is used. The volume fraction values are linearly interpolated as given by the following expression:

\begin{equation}
V(x,y) = \boldsymbol{N \tilde{V_{e}^{xy}}},
\label{interpolation_eq}
\end{equation}

where \(x\in[x_{i},x_{i+1}]\),   \(y\in[y_{j},y_{j+1}]\), $\boldsymbol{\tilde{V}^{xy}_{e}}$ is given by  $[V^{xy}_{i,j},\;V^{xy}_{i+1,j},\; V^{xy}_{i+1,j+1},\; V^{xy}_{i,j+1}]$ and linear interpolation functions $\boldsymbol{N}$ is given by:
\begin{equation}
\begin{aligned}
\boldsymbol{N} =  \left[\frac{(1-\xi)(1-\eta)}{4}, \quad \frac{(1+\xi)(1-\eta)}{4}, \right. \\
 \left. \frac{(1+\xi)(1+\eta)}{4}, \quad \frac{(1-\xi)(1+\eta)}{4} \right],    
\end{aligned}
\end{equation}

where $\xi, \eta \in [-1, 1]$ are parametric coordinates within an element.

\section{Genetic Algorithm} \label{sec_GA}

In this manuscript, we employ the genetic algorithm to find the optimum volume fraction distribution of FGMs. GA is a derivative-free optimization algorithm \citep{goldberg1989genetic} and has been shown to work robustly for a large number of applications. Although in general the overall methodology of the GA is quite well established in the literature, there is no existing work where GA has been used in conjunction with the design space generated by GPR. In this section, we provide an overview of the GA algorithm, with emphasis on the modifications required to make it compatible with GPR-based design space. The overall genetic process consists mainly of three key steps: crossover, mutation, and tournament selection. 
\subsection{Modified Crossover scheme}
We have used the Simulated Binary (SBX) crossover operator \citep{deb1995simulated} in our implementation of the FGM optimization scheme. In standard SBX, children are generated as a combination of the parents. With parent vectors $\boldsymbol{P}_1$ and $\boldsymbol{P}_2$, the volume fraction for children  $\boldsymbol{C}_1$ and $\boldsymbol{C}_2$ at each node can be straightforwardly calculated using the expressions given below: 

\begin{equation}
\begin{aligned}
\boldsymbol{C}_1 &= 0.5\left[(1+\boldsymbol{\beta})\odot \boldsymbol{P}_1 + (1-\boldsymbol{\beta})\odot \boldsymbol{P}_2\right],\\
\boldsymbol{C}_2 &= 0.5\left[(1-\boldsymbol{\beta})\odot \boldsymbol{P}_1 + (1+\boldsymbol{\beta})\odot \boldsymbol{P}_2\right],
\end{aligned}
\label{cross_eq}
\end{equation}
where, $\odot$ represents the node-wise  product, and $\boldsymbol{\beta}$ is the vector of spread factors $\beta_i$ given by the following expressions.

\[
\beta_i =
\begin{cases}
(2r_i)^{\frac{1}{\eta + 1}}, & \text{if } r_i \leq 0.5, \\
\left(\dfrac{1}{2(1 - r_i)}\right)^{\frac{1}{\eta + 1}}, & \text{if } r_i > 0.5.
\end{cases}
\]
Here, $\eta$ denotes the crossover strength parameter that governs the shape of the spread, and $\boldsymbol{r}$ is a random vector with each component $r_i \in [0, 1]$.

\begin{figure*}
	\centering
		\includegraphics[scale=0.60]{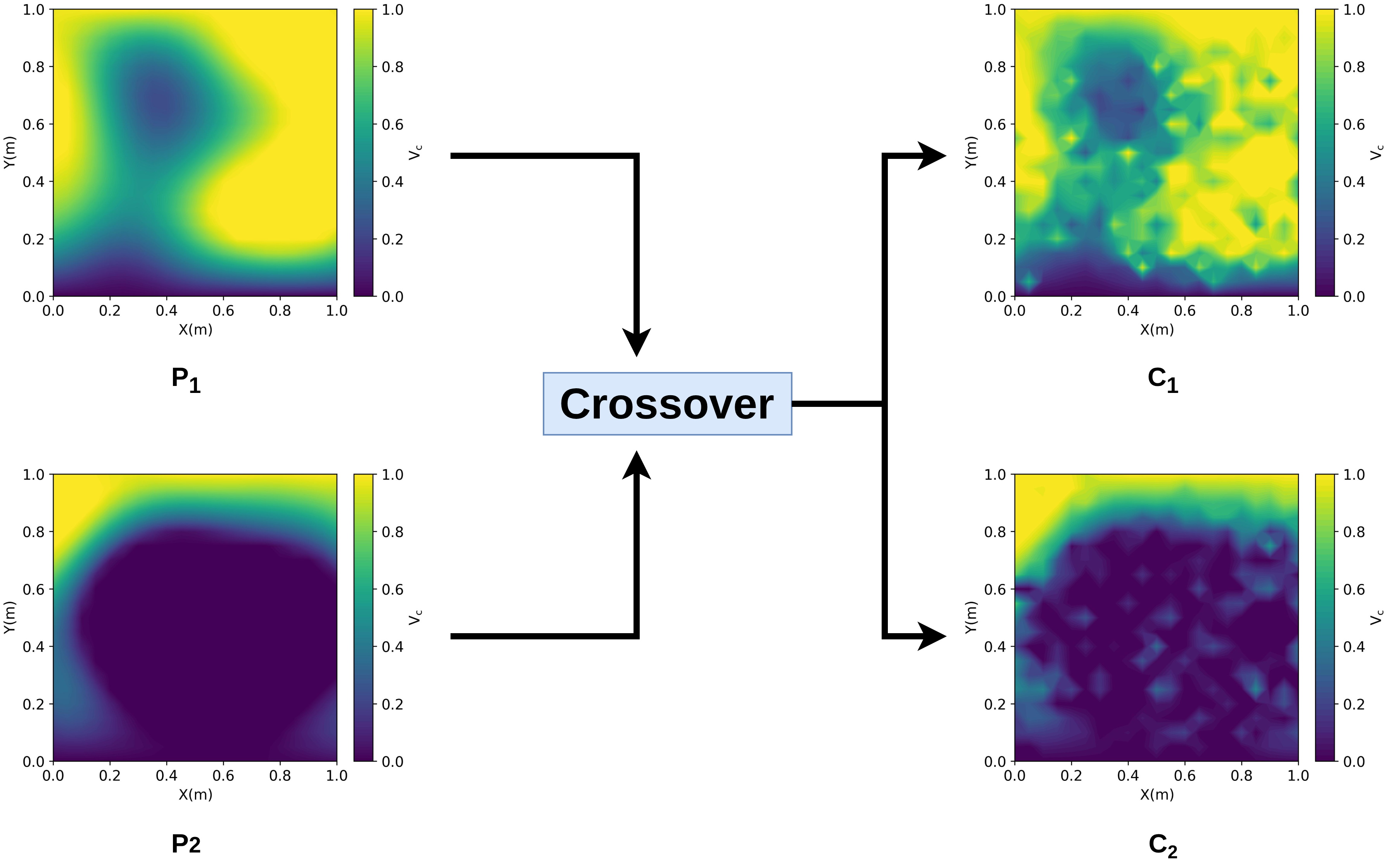}
	\caption{Comparison of FGM profiles before ($P_1$, $P_2$) and after crossover ($C_1$, $C_2$) operation, highlighting the reduction in the smoothness of the FGM design.}
	\label{FIG_Crossover}
\end{figure*}

However, as illustrated in Fig.~\ref{FIG_Crossover}, the offspring produced through the combinations given by Eq.~\ref {cross_eq} exhibit a disruption in the smoothness of the FGM profile. These non-smooth profiles might lead to stress concentration, and thus are not desirable from a practical point of view. In terms of the underlying multivariate distribution (Eq.~\eqref{eq:prior_dist}), the likelihood of these non-smooth profiles is minimal. Hence, these non-smooth profiles do not fall into the underlying design space. 

In order to obtain the smooth profiles, we project the non-smooth profiles into the underlying design space. To do so, we carry out the projection using the following expression:
\begin{equation} 
\label{proj}
    \boldsymbol{C}^{'}=\boldsymbol{K} (\boldsymbol{K}+\sigma^2 \boldsymbol{I})^{-1}\boldsymbol{C}.
\end{equation}

\begin{figure}[ht]
	\centering
		\includegraphics[scale=0.60]{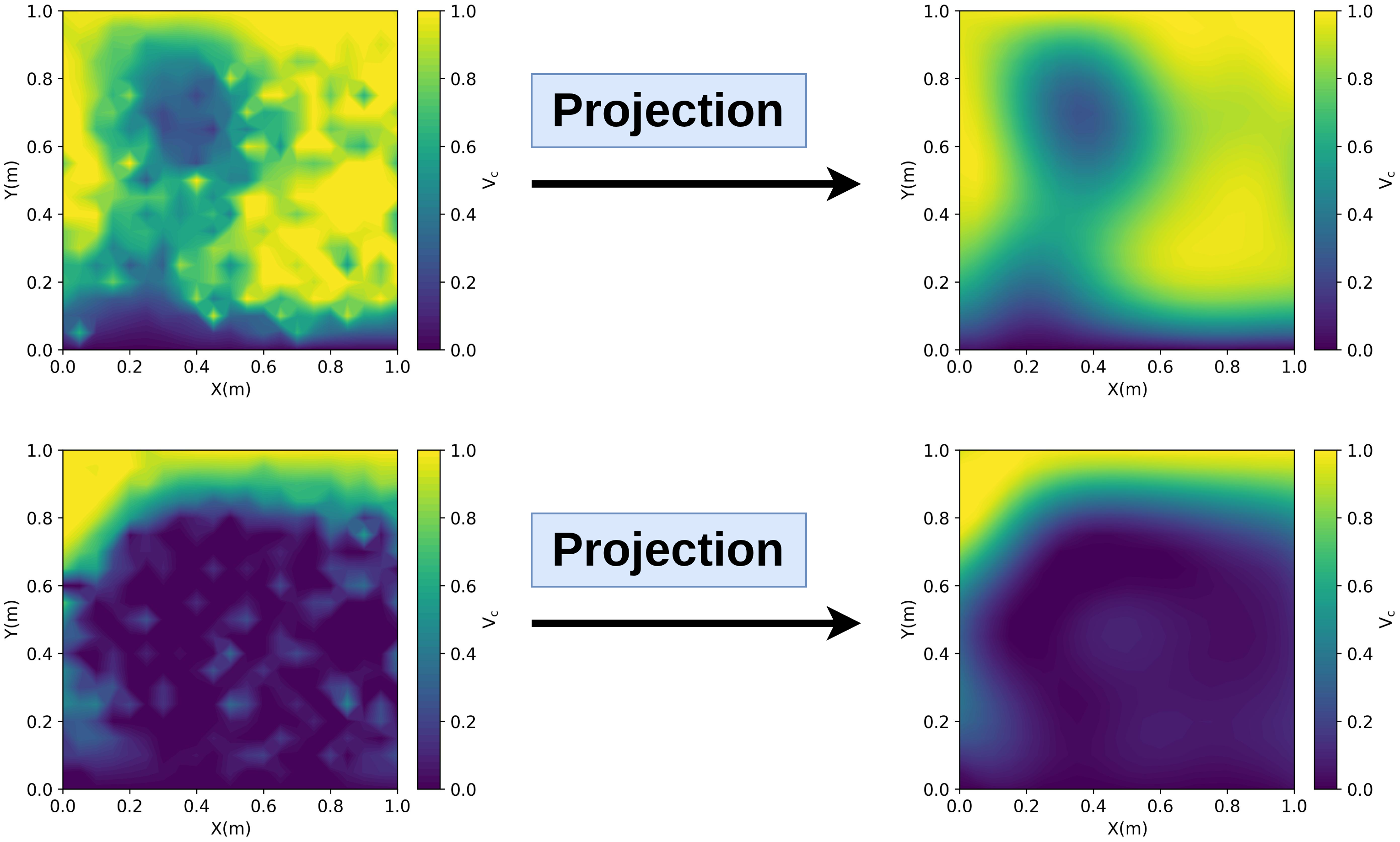}
	\caption{Projection of the FGM profiles obtained after the crossover operation into a smoother design space.}
	\label{FIG_Child_GPR}
\end{figure}

Here,$\boldsymbol{C}^{'}$ is the projected vector, \({\boldsymbol{K}}\) is the same covariance matrix as used in the design space generation and can be calculated from the radial basis kernel function, as stated in Eq.~\eqref{RBF}. The hyperparameters \(l\) and \(\sigma\) can be taken same as the ones used for the generation of the underlying design space. One example of the projection operation has been shown in Fig.~\ref{FIG_Child_GPR}, where the projection operator has been applied to two of the profiles obtained by the SBX crossover operator. As can be seen from the Fig.~\ref{FIG_Child_GPR}, the above projection results in the smoothening of the profiles, while retaining the underlying characteristics.

\begin{figure*}
    \centering
    \begin{subfigure}{0.22\textwidth}
        \centering
        \includegraphics[width=\linewidth]{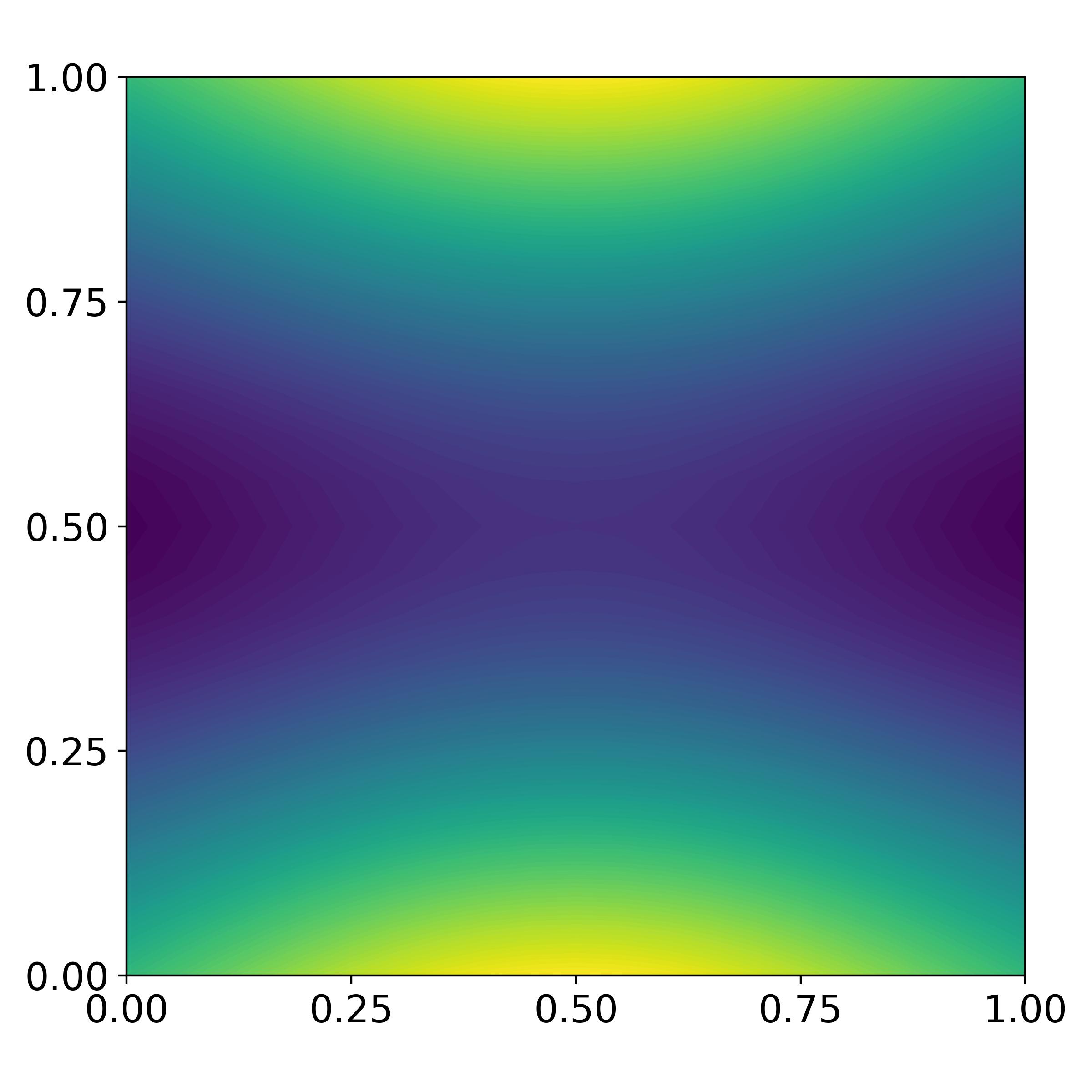}
        \caption{}
    \end{subfigure}
    \begin{subfigure}{0.22\textwidth}
        \centering
        \includegraphics[width=\linewidth]{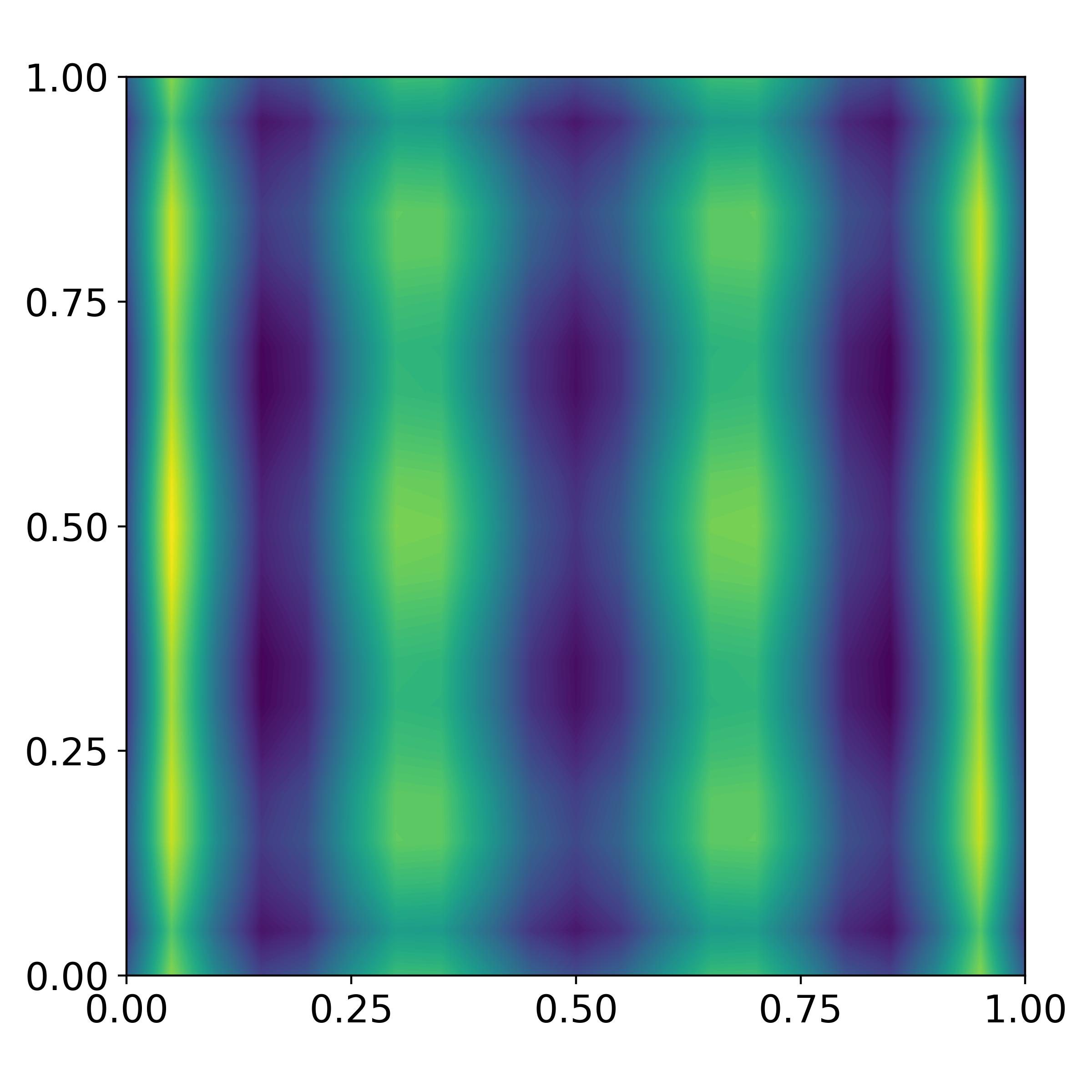}
        \caption{}
    \end{subfigure}
    \begin{subfigure}{0.22\textwidth}
        \centering
        \includegraphics[width=\linewidth]{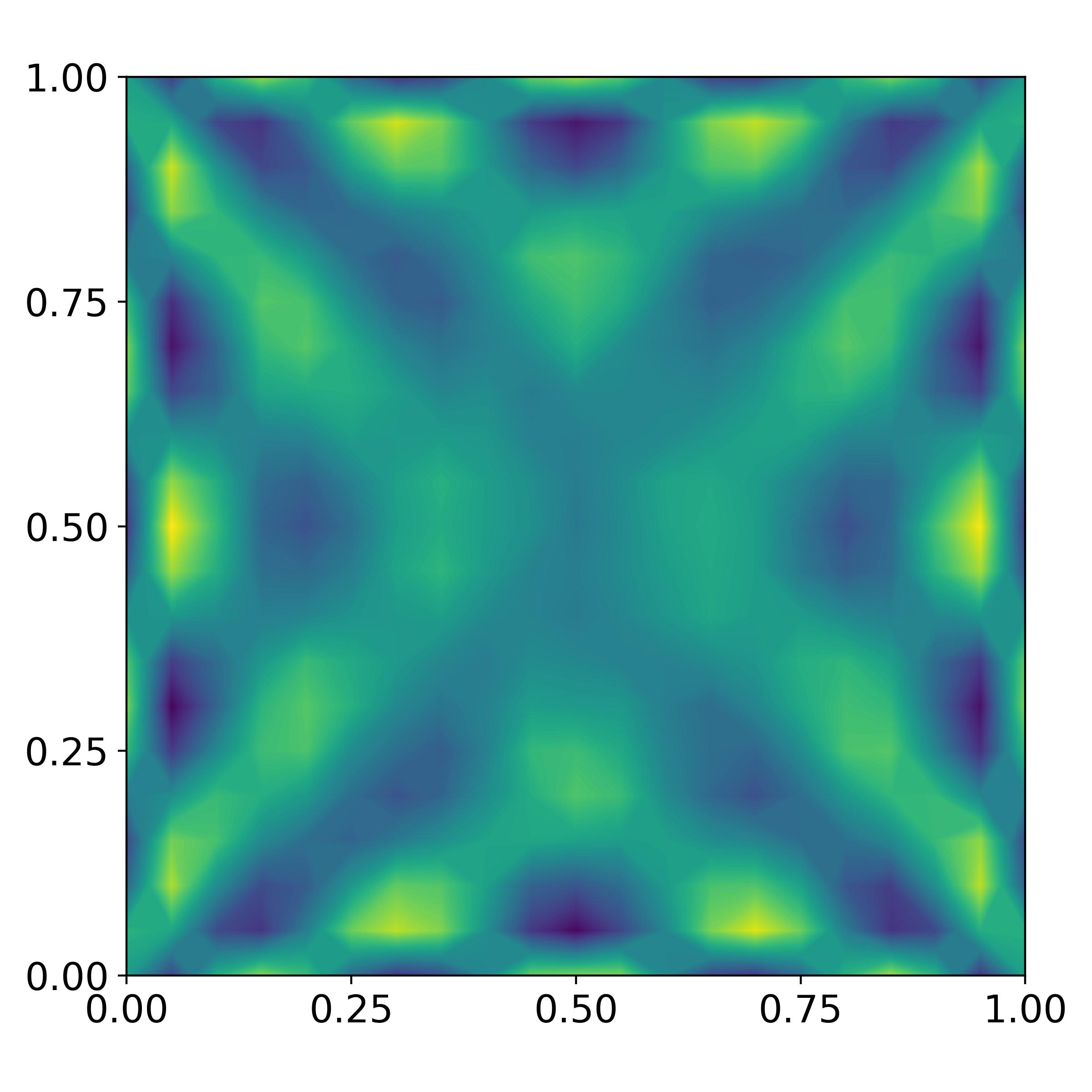}
        \caption{}
    \end{subfigure}
    \begin{subfigure}{0.22\textwidth}
        \centering
        \includegraphics[width=\linewidth]{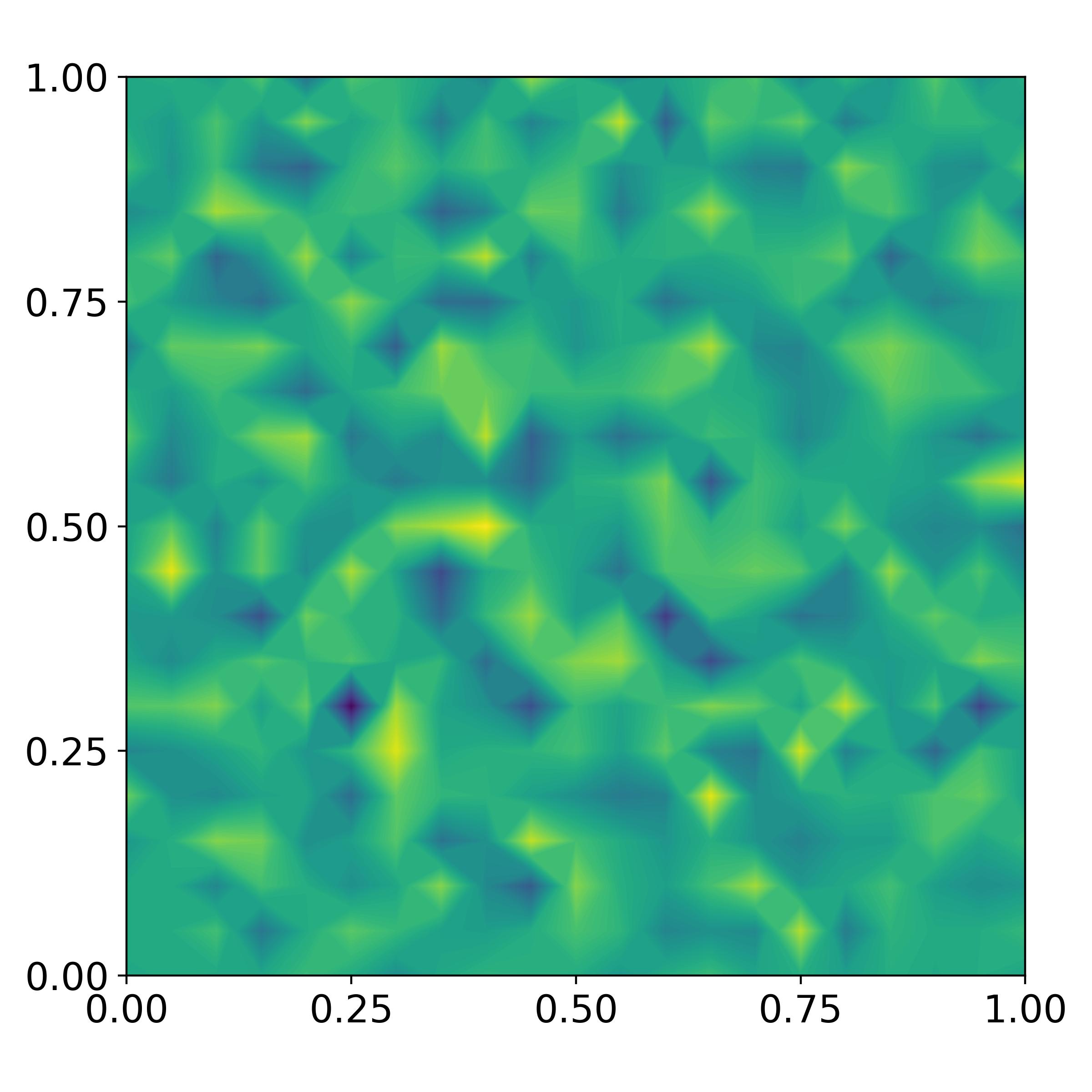}
        \caption{}
    \end{subfigure}
    
    \caption{Eigenvectors corresponding to the (a) $5^{th}$, (b) $50^{th}$, (c) $100^{th}$, and (d) $200^{th}$ largest eigenvalue of the covariance matrix. This illustrates that the smoother eigenvector profiles are obtained for the higher eigenvalues.}
    \label{eigenvectors}
\end{figure*}

To understand the working of the above projection operator, consider that \(\lambda_1,\lambda2,...\) and \(\boldsymbol{e_1},\boldsymbol{e_2}\) denote the eigenvalues and eigenvectors of the covariance matrix \(\boldsymbol{K}\). Without any loss of generality, we can assume that the eigenvalues are arranged in decreasing order. Now, considering that eigenvectors from an orthonormal basis, the vector \(\boldsymbol{C}\) can be written as:
\begin{equation*}
    \boldsymbol{C}=\alpha_1\boldsymbol{e_1}+\alpha_2\boldsymbol{e_2}+...+\alpha_{n+m}\boldsymbol{e_{n+m}}.
\end{equation*}
Thus, we can write Eq.~\eqref{proj} as :
\begin{align*}
\boldsymbol{C}^{'}=\boldsymbol{K} (\boldsymbol{K}+\sigma^2 I)^{-1}\boldsymbol{C} =&\frac{\alpha_1\lambda_1}{\lambda_1+\sigma^2}\boldsymbol{e_1}+\frac{\alpha_2\lambda_2}{\lambda_2+\sigma^2}\boldsymbol{e_2}+...+\frac{\alpha_{n+m}\lambda_{n+m}}{\lambda_{n+m}+\sigma^2}\boldsymbol{e_{n+m}}\\
=& \frac{\alpha_1}{1+\frac{\sigma^2}{\lambda_1}}\boldsymbol{e_1}+\frac{\alpha_2}{1+\frac{\sigma}{\lambda_2}}\boldsymbol{e_2}+...+\frac{\alpha_{n+m}}{1+\frac{\sigma^2}{\lambda_{n+m}}}\boldsymbol{e_{n+m}}.
\end{align*}
Now, consider a number \(p\), such that \(\lambda_{p+j}<<\sigma^2\), where \(j=1,2..n+m-p\), leading to \(\frac{\alpha_{p}}{1+\frac{\sigma^2}{\lambda_{p}}} \approx 0\). Thus, we can write, the  \(\boldsymbol{C}^{'}\) as:
\begin{equation*}
    \boldsymbol{C}^{'}\approx \frac{\alpha_1}{1+\frac{\sigma^2}{\lambda_1}}\boldsymbol{e_1}+\frac{\alpha_2}{1+\frac{\sigma}{\lambda_2}}\boldsymbol{e_2}+...+\frac{\alpha_{p}}{1+\frac{\sigma^2}{\lambda_{p}}}\boldsymbol{e_{p}}.
\end{equation*}
Thus, the projected vector \(\boldsymbol{C}^{'}\) only contains a few leading eigenvectors. The eigenvectors associated with large eigenvalues are smooth in nature, while as the eigenvalue decreases, the smoothness of the corresponding eigenvector decreases. The same has been shown in the Fig. \ref{eigenvectors}. Since the projected vector contains only the leading eigenvectors, which are smooth in nature, the projected vector is also smooth in nature.

\subsection{Mutation}The next step in the evolutionary process is mutation, where perturbations are introduced to the offspring generated during crossover. In this work, we employ a Gaussian mutation operator. The perturbation values are sampled from a multivariate Gaussian distribution with zero mean as follows.
\begin{equation}
\begin{aligned}
    \boldsymbol{p} &\sim \mathcal{N}\left(\boldsymbol{0},\boldsymbol{K^{\star}} \right).
\end{aligned}
\label{eq:f_star_given_f}
\end{equation}

The covariance matrix is given by Eqn. \eqref{K_exp}, and is identical to the one used during design space generation. Note, at the boundary points where volume fraction has been specified, variance is close to zero. Further, since the underlying mean is also zero, the perturbation at these nodes is close to zero. The sampled perturbation profile is then scaled by a factor \(\epsilon\) and thus the designs after the mutation \(\boldsymbol{C_m}^{'}\)are calculated using the below expression.
\begin{equation}
    \boldsymbol{C_m}^{'}=\boldsymbol{C}^{'}+\epsilon \boldsymbol{p}.
\end{equation}

Note in the numerical examples, the value of the \(\epsilon\) is taken in the range of \(0.2-0.3\). The consistency of the mutation operator with the profile generation algorithm ensures smoothness of the FGM profiles after mutation. 

\subsection{Tournament selection}

For the evolution of the population, this study employs tournament selection as the selection operator, in which a small subgroup of FGM profiles is randomly chosen from the population. The FGM profile with the highest fitness score within each selected subgroup proceeds to the next stage of the evolution. FEA has been used as the underlying technique to compute the fitness value of each profile. Further, we incorporate an elitism strategy to preserve the best profile from each generation. In this strategy, the single best profile identified during the fitness evaluation is retained before the tournament selection and used to replace the worst profile after the mutation.

\subsection{Constraint handling in genetic algorithm}
In the literature, various methods are available to handle the inequality constraints in GA \citep{datta2015evolutionary}, such as penalty function, repair algorithm, decoder, and hybrid approach. In our work, we use the constraint handling technique proposed by K.Deb \citep{deb2000efficient}, which is free from any penalty parameter. The constrained optimization problem is written in the form:

\begin{equation}
\begin{aligned}
\textbf{Minimize:}\; & \quad f(V_{c}^{i}), \quad  i =1,2,..,n,\\ 
\textbf{Subject to:}\; & \quad g_{k}(V_{c}^{i})  \le g_{k}^{*}, \quad k =1,2,..,K,
\end{aligned}
\end{equation}
Here, $n$ denotes the total number of nodes, $V_{c}^{i}$ represents the volume fraction of ceramic at the $i^{th}$ node, $f(V_{c}^{i})$ is the objective function, and $g_{k}$ denotes the $k^{th}$ inequality constraint. The fitness function $F(x)$ is then constructed according to the following equations:

\begin{equation}
\begin{aligned}
F(V^{(i,j)}) =
\begin{cases}
    f(V_{c}^{i}) & \text{if }\quad g_{k}(V_{c}^{i})  \le g_{k}^{*}, \quad k =1,2,..,K, \\
    f_{\max} + \sum\limits_{k=1}^{K} |\langle g_k(V_{c}^{i}) \rangle| &   \text{otherwise},
\end{cases}
\end{aligned}
\end{equation}
\noindent where, $f_{\max}$ is worst feasible solutions.

\section{Numerical examples} \label{sec_NE}
In this section, we present various numerical examples to demonstrate the efficacy of the proposed optimization framework for FGM. To demonstrate the efficacy of the framework comprehensively,  we present both unconstrained and constrained problems. The examples are arranged in increasing order of the geometric complexity of the underlying domain. In all the examples, we have considered the case of the thermo-elastic loading. In all the examples, we consider the FGM consisting of Aluminum (Al) as the metallic phase and zirconia (ZrO\textsubscript{2}) as the ceramic phase. The material properties used for Al and ZrO\textsubscript{2} are listed in Table \ref{tab:prop_al_zro2}. The proposed methodology incorporates a GPR-based algorithm to generate the FGM profiles, as described in Section~\ref{PGA}, thermoelastic finite element analysis to evaluate the maximum von Mises stress ($\sigma^{v}_{max}$) and the temperature field, as detailed in Section~\ref{FEM}, and optimization using genetic algorithm as mentioned in the Section \ref{sec_GA}.

\subsection{Problem 1: Rectangular FGM plate subjected to sinusoidal temperature loading}
\begin{figure}[htbp]
  \centering
  \includegraphics[width=0.55\textwidth]{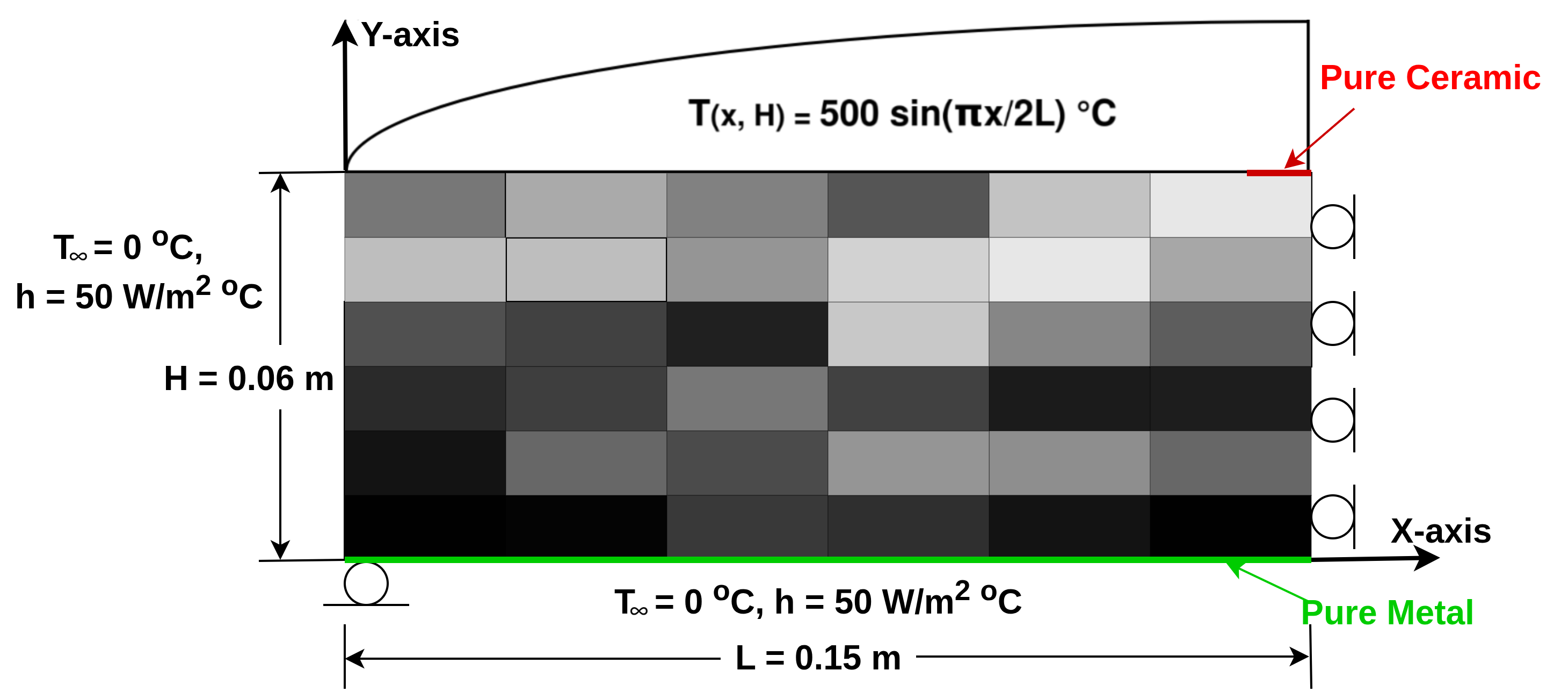}
  \caption{Problem 1: Schematic of rectangular $\mathrm{Al} / \mathrm{ZrO_2}$ FGM plate subjected to the sinusoidal temperature loading.}
  \label{prb1_diagram}
\end{figure}

In the first numerical example, we consider the problem of a simply supported FGM plate under plane stress conditions as shown in Fig. \ref{prb1_diagram}, similar to the problem studied by Goupee et al. \citep{goupee2006two}. We intend to minimize the $\sigma_{max}^{v}$ for one half of the domain. The top edge of the plate is subjected to a sinusoidal temperature $\theta = 500 \sin\!\left(\tfrac{\pi x}{2L}\right)\,^{\circ}\mathrm{C}$ condition. While the right edge belongs to a symmetry boundary condition with displacement along the X-axis is zero ($u_1$ = 0) and adiabatic condition ($q = 0$). Left and bottom edge is subjected to convective heat transfer condition with the convection heat transfer coefficient, $h$ = \SI{50}{\watt\per\metre\squared\per\degreeCelsius} and the surrounding temperature is maintained at $\theta_{\infty}$ = \SI{0}{\degreeCelsius}. The bottom left corner is subjected to zero displacement along the Y-axis. In the design domain, pure Al ($V_{c} = 0$) is prescribed along the bottom edge, while pure ZrO$_2$ ($V_{c} = 1$) is enforced at the top right edge, limited to 10\% of its length. This choice is made due to the  ZrO$_2$ superior thermal properties.

\begin{table}
\centering
\begin{minipage}{0.40\textwidth}
    \centering
    \caption{Material properties of aluminum and zirconia.}
    \begin{tabular}{p{3cm}p{1.0cm}p{1.0cm}}
\hline 
\textbf{Material Property} & \textbf{Al}  & \textbf{ZrO$_2$} \\ 
\hline 
$E$(GPa)   & 70.0   & 200.0   \\ 
$\nu $  & 0.3   & 0.3   \\ 
$\alpha(10^{-6} \mathrm{\kappa}^{-1})$   & 23.4   & 10.0  \\
$k$ (W/mK)   & 233.0   & 2.2   \\
$\rho$ (kg/$\text{m}^{3}$)   & 2707   & 5700   \\ 
\hline 
    \end{tabular}
    \label{tab:prop_al_zro2}
\end{minipage}
\hspace{0.3 cm}
\begin{minipage}{0.40\textwidth}

    \centering
    \caption{Parameters of the genetic algorithm.}
    \begin{tabular}{p{3.0cm}p{4.0cm}}
    \hline
        \textbf{Parameter} & \textbf{Value} \\
        \hline
        Population size & 200 \\
        Tournament size & 4 \\
        Crossover parameter & $\eta = 1.5\left[1+\tfrac{1}{2}\left(1-e^{-g/100}\right)\right]$ \\
        Mutation parameters & $l = 0.05$, $\sigma = 1$ \\
        Mutation probability & 0.3 \\
        \hline
    \end{tabular}
    \label{tab:P1_GA_Para}
\end{minipage}
\end{table}

\begin{figure}[h!]
    \centering
    
    \begin{subfigure}{\textwidth}
        \centering
        \begin{subfigure}{0.9\textwidth}
            \includegraphics[trim=0 50 0 30,clip, width=\linewidth]{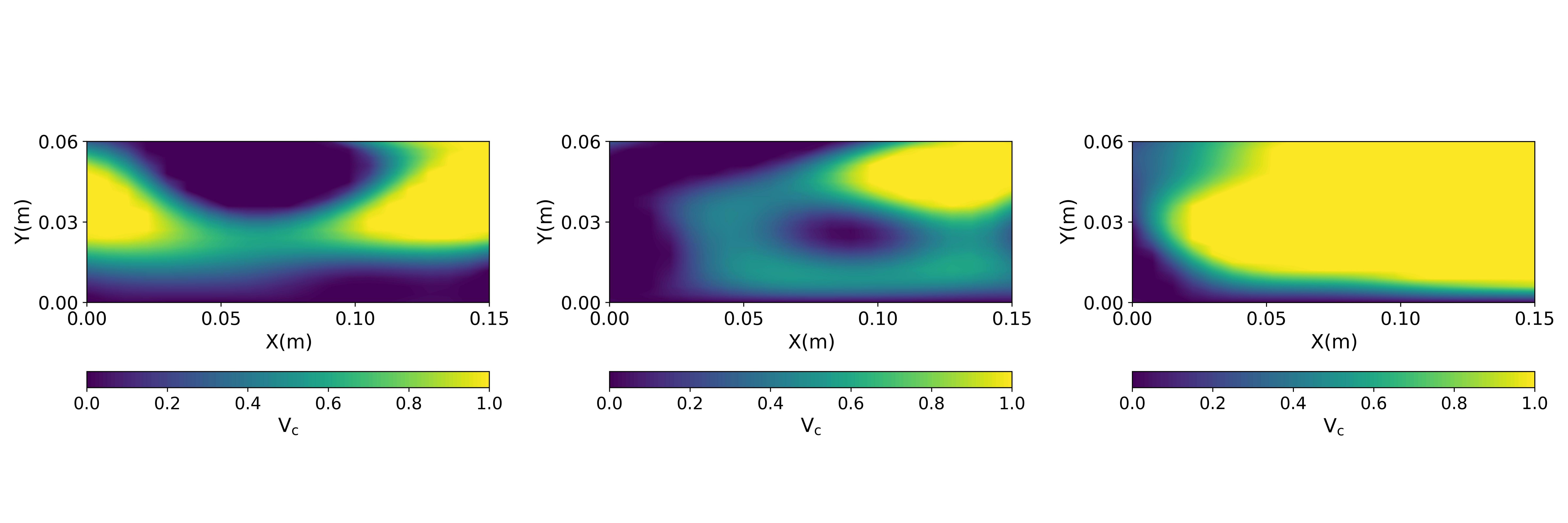}
        \end{subfigure}
        \caption{}
        \label{random_prob1_vof}
    \end{subfigure}
    
    \vspace{0.1em} 
    
    \begin{subfigure}{\textwidth}
        \centering
        \begin{subfigure}{0.9\textwidth}
            \includegraphics[trim=0 60 0 50,clip,width=\linewidth]{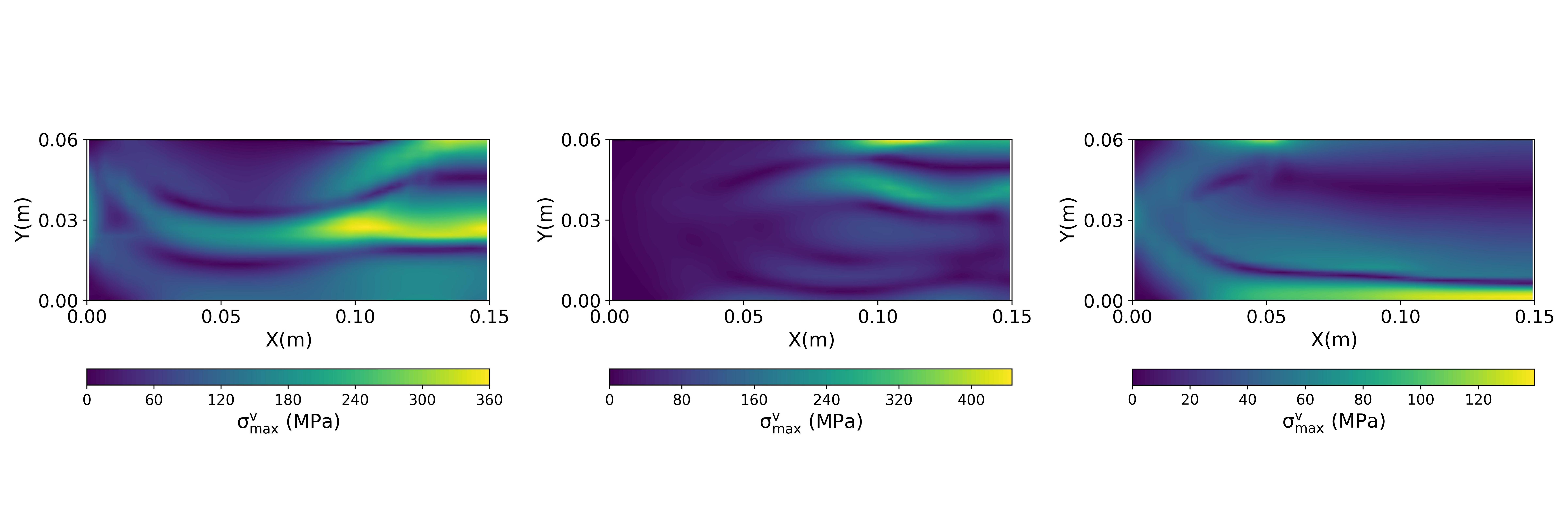}
        \end{subfigure}
        \caption{}
        \label{random_prob1_stress}
    \end{subfigure}

    \vspace{0.1em} 
    
    \begin{subfigure}{\textwidth}
        \centering
        \begin{subfigure}{0.9\textwidth}
            \includegraphics[trim=0 60 0 50,clip,width=\linewidth]{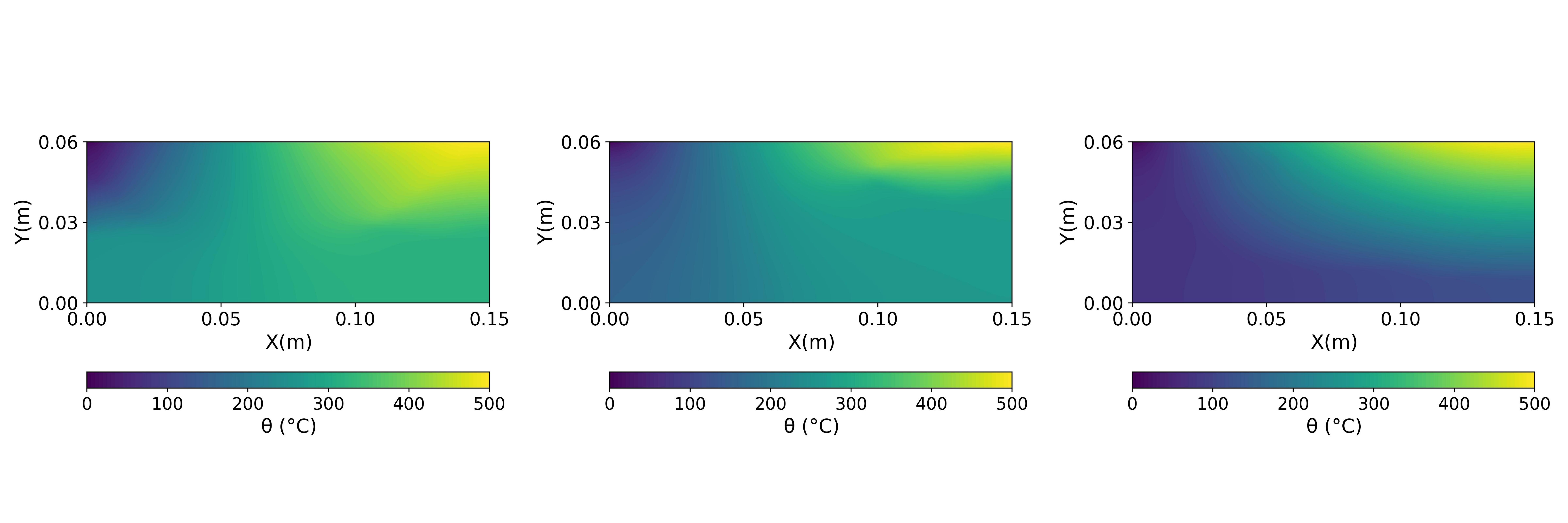}
        \end{subfigure}
        \caption{}
        \label{random_prob1_temp}
    \end{subfigure}
    
    \caption{Sample FGM profiles for the rectangular plate generated through GPR-based profile generation algorithm: (a) contour plot of the ceramic volume fraction distribution, (b) corresponding von Mises stress distribution, and (c) temperature distribution. }
    \label{random_prof_prob1}
\end{figure}

 We consider two different cases with the objective to minimize the $\sigma_{max}^{v}$, of unconstrained and constrained optimization respectively. GA parameters for both of them are listed in Table \ref{tab:P1_GA_Para}. For FEA, we discretize our domain into 9 noded quadrilateral elements with a mesh density of $20\times20$ elements. The same discretization has been considered to define the volume fraction profile.  The termination criteria for the GA are defined as follows: (1) a minimum of 100 generations must be completed, and (2) the improvement in the fitness of the best individual must not exceed 0.1 MPa over the last 10 generations. The parameters of GPR to generate the FGM profiles are $l$ = 0.05 and $\sigma$ = 1.0. 
 Three random FGM profiles generated through the proposed algorithm are shown in Fig. \ref{random_prob1_vof}, and the corresponding von Mises stress and temperature distribution are shown in Figs. \ref{random_prob1_stress} and \ref{random_prob1_temp}, respectively. As can be seen from the figure, the generated profiles are smooth in nature and adhere to the specified volume fraction at the part of the boundaries. However, we can observe the spike in the underlying stress distributions for all the three profiles.

\subsubsection{Case 1}
In the first case, we perform the unconstrained optimization to minimize the $\sigma_{max}^{v}$ to find the optimal volume fraction distribution in the simply supported  FGM plate. The optimization problem is stated as follows:
\begin{equation}
\textbf{Minimize}: \quad \sigma_{max}^{v}(V^i_{c}), \quad i=1,2,..,n. 
\end{equation}

\begin{figure*}[ht]
    \centering
    \begin{subfigure}[b]{0.33\textwidth}
        \centering
        \includegraphics[width=\textwidth]{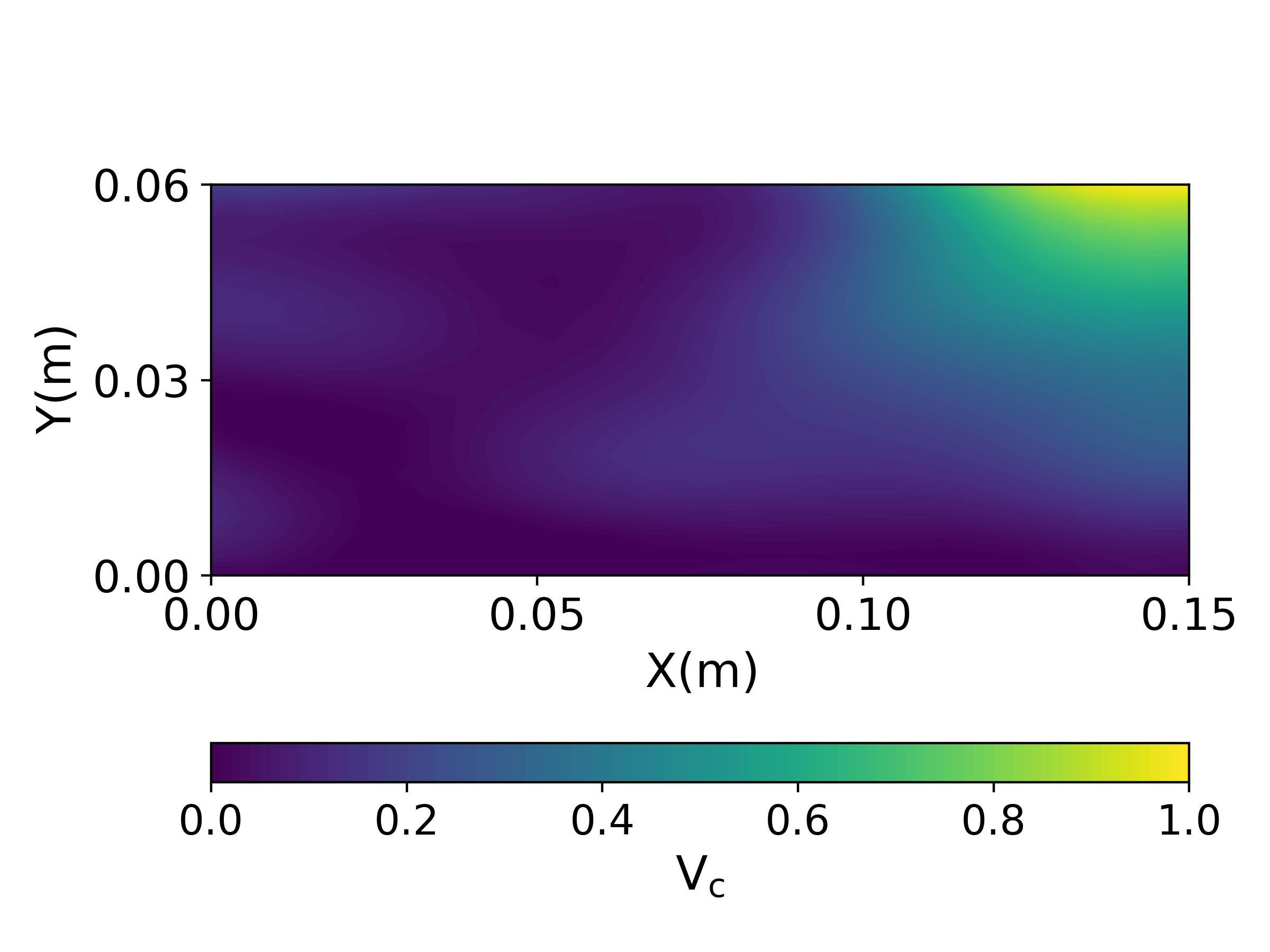}
        \caption{}
        \label{vof_1_1a}
    \end{subfigure}
    \begin{subfigure}[b]{0.33\textwidth}
        \centering
        \includegraphics[width=\textwidth]{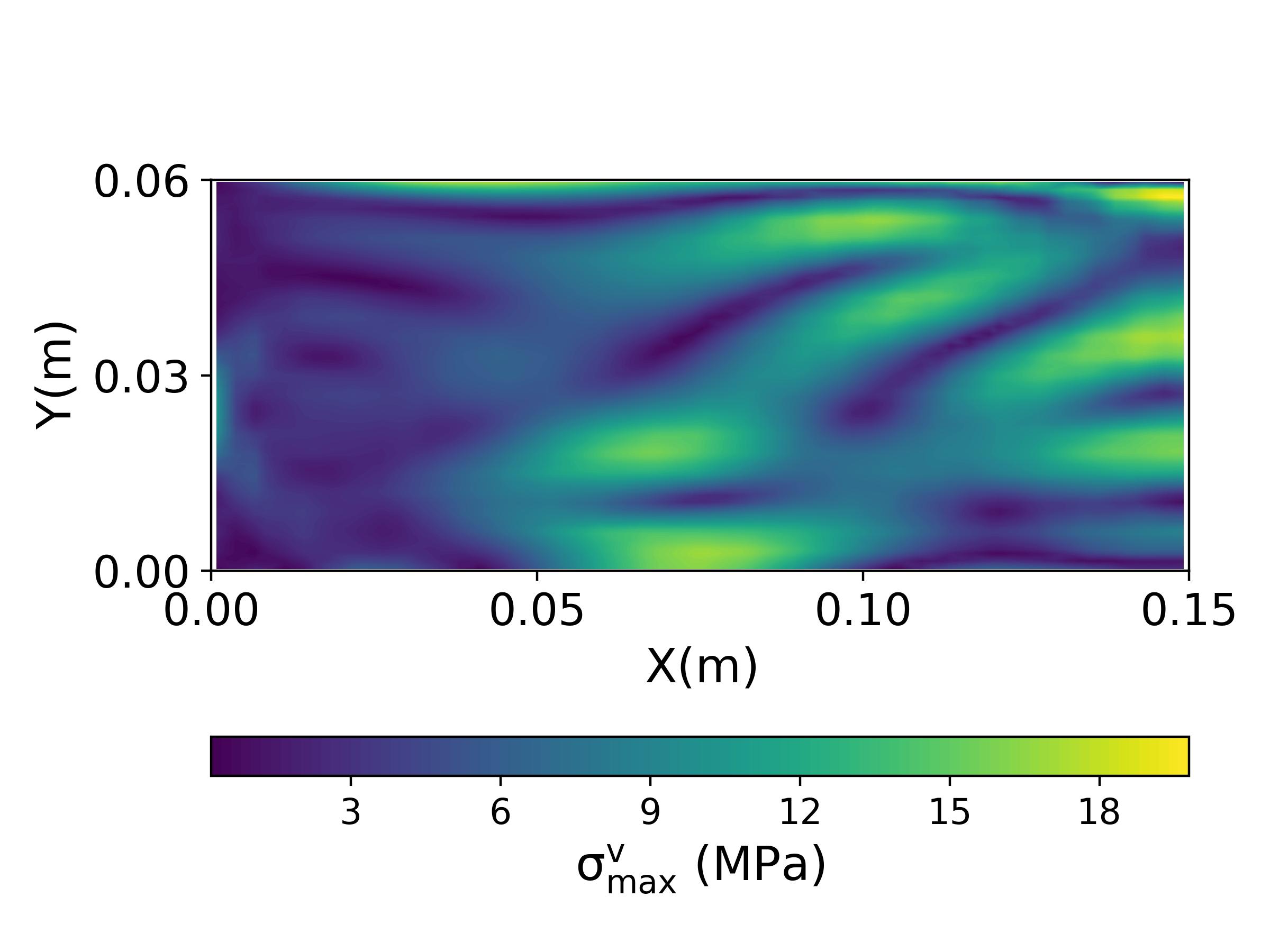}
        \caption{}
        \label{stress_1_1a}
    \end{subfigure}
    \begin{subfigure}[b]{0.33\textwidth}
        \centering
        \includegraphics[width=\textwidth]{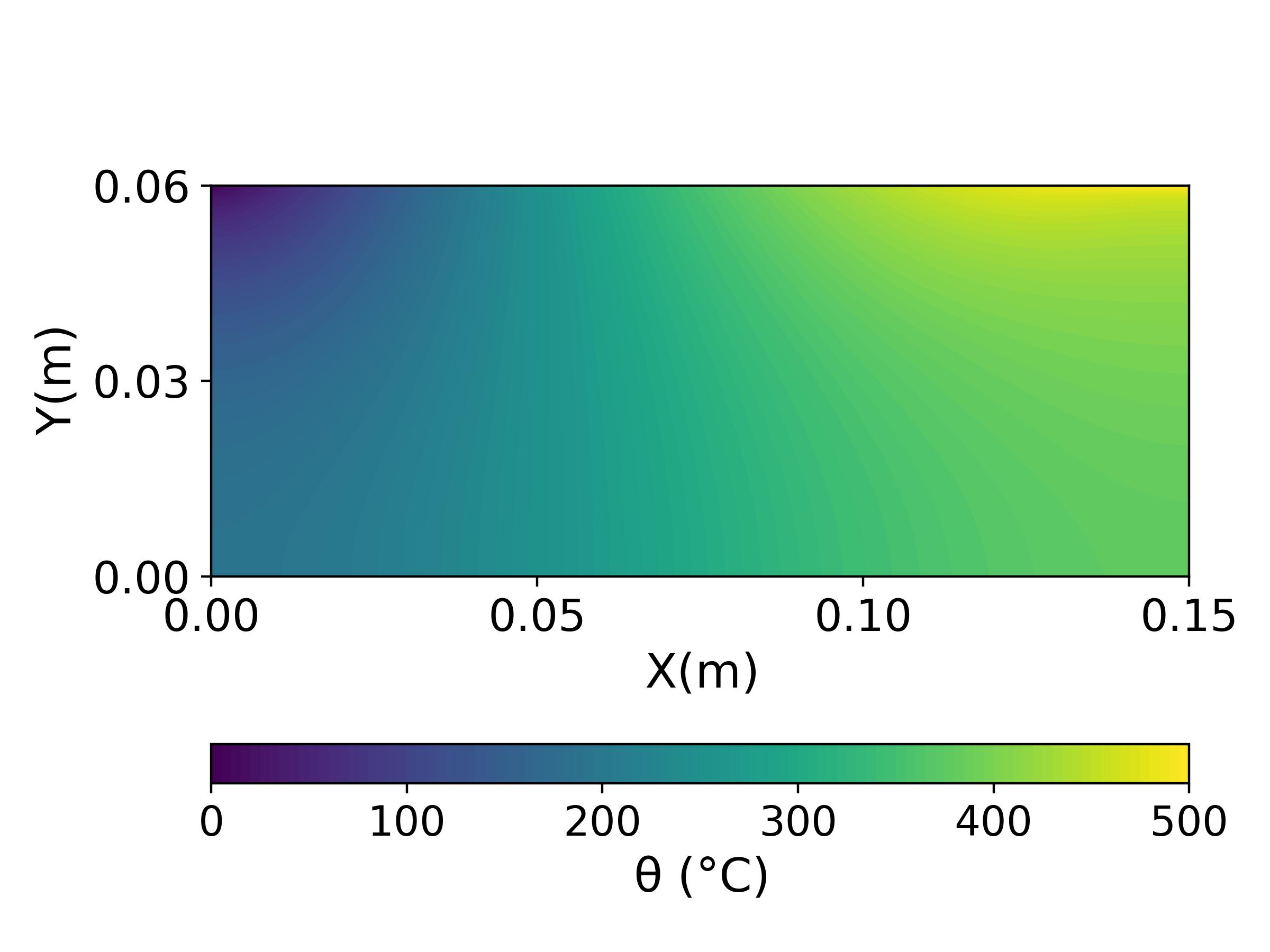}
        \caption{}
        \label{temp_1_1a}
    \end{subfigure}
    \caption{Optimized FGM plate subjected to the sinusoidal temperature loading at the top surface: (a) contour plot of the ceramic volume fraction distribution, (b) corresponding von Mises stress distribution, and (c) temperature distribution.}
    \label{}
\end{figure*}

\begin{figure}[htbp]
  \centering
  \includegraphics[width=0.37\textwidth]{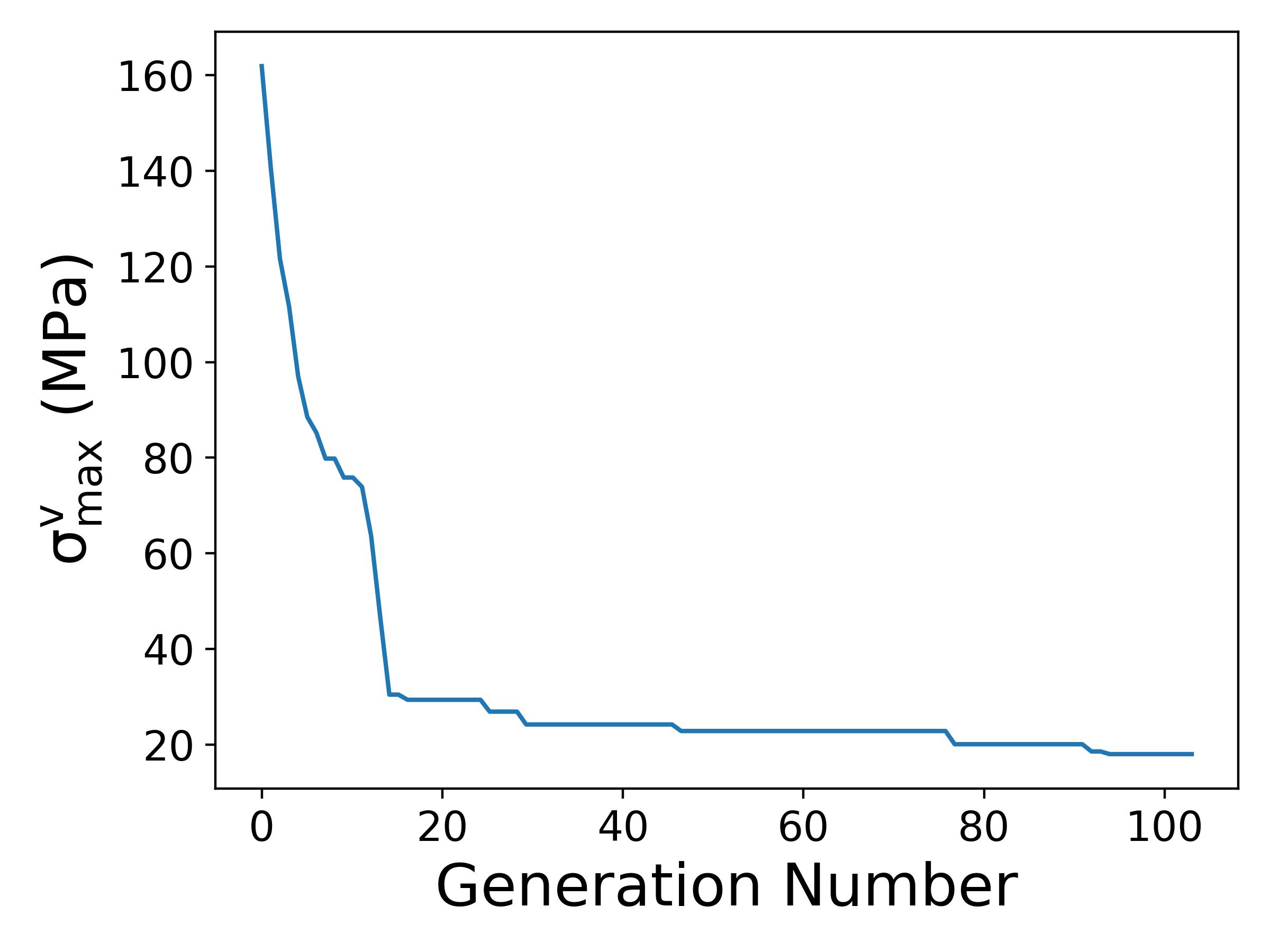}
  \caption{Evolution of the best fitness value over generations during the optimization of a rectangular FGM plate subjected to the sinusoidal temperature loading conditions.}
  \label{conv_1a}
\end{figure}

The volume fraction distribution of the optimized FGM profile is shown in Fig. \ref{vof_1_1a} and the evolution of profiles with respect to generations is shown in Fig. \ref{conv_1a}. The figure shows that the optimized FGM profile successfully adheres to the prescribed volume fraction on part of the boundaries. As can also be seen from the figure, the optimum profile has a smooth gradation and ensures a smooth transition from the pure ceramic region to the pure metal region. More importantly, as can be seen from  Fig. \ref{stress_1_1a}, the stress variation for the optimum profile is significantly smooth in nature compared to the random profiles (Fig. \ref{random_prob1_stress}), and the regions of stress concentration are absent. The optimal value obtained for $\sigma_{max}^{v}$ is 19.4 MPa. For the comparison with the FGM profile having the linear gradation of volume fraction along the Y-axis, the $\sigma_{max}^{v}$ is 131.6 MPa, and for the profile having bi-linear volume fraction gradation, $\sigma_{max}^{v}$ is 158.5 MPa.\\

\subsubsection{Case 2}
In the second case, a constrained optimization is performed to minimize the $\sigma_{max}^{v}$ and determine the optimal volume fraction distribution in the FGM plate. The constraint imposed is that the volume fraction of ceramic must exceed a prescribed threshold ($V_{c}^{*}$) at locations where the temperature surpasses a critical value ($\theta^{*}$). This constraint is practically motivated, as ceramics possess higher thermal resistance than metals, making such a distribution more suitable for thermo-mechanical loading conditions. The optimization problem is stated as follows:

\begin{equation}
\begin{aligned}
&\textbf{Minimize:} \quad && \sigma_{max}^{v}(V^i_{c}), \quad i = 1,2,\dots,n, \\
&\textbf{Subject to:} \quad && V_{c}(x) \ge V_{c}^{*}, \quad \text{if } \theta(x) \ge \theta^{*}, \quad x \in \Omega.
\end{aligned}
\end{equation}

\begin{figure*}[ht]
    \centering
    \begin{subfigure}[b]{0.33\textwidth}
        \centering
        \includegraphics[width=\textwidth]{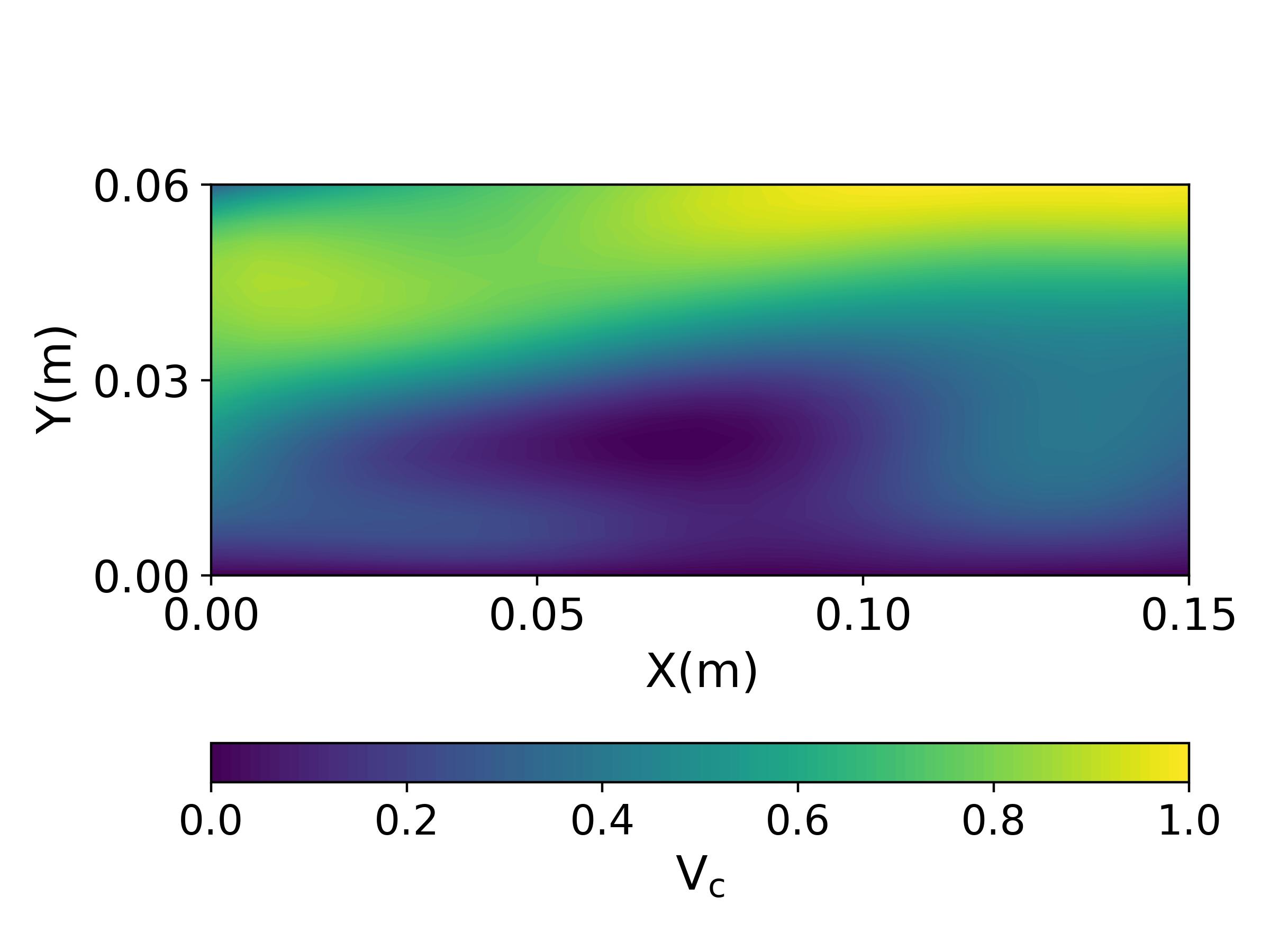}
        \caption{}
        \label{vof_1_2a}
    \end{subfigure}
    \begin{subfigure}[b]{0.33\textwidth}
        \centering
        \includegraphics[width=\textwidth]{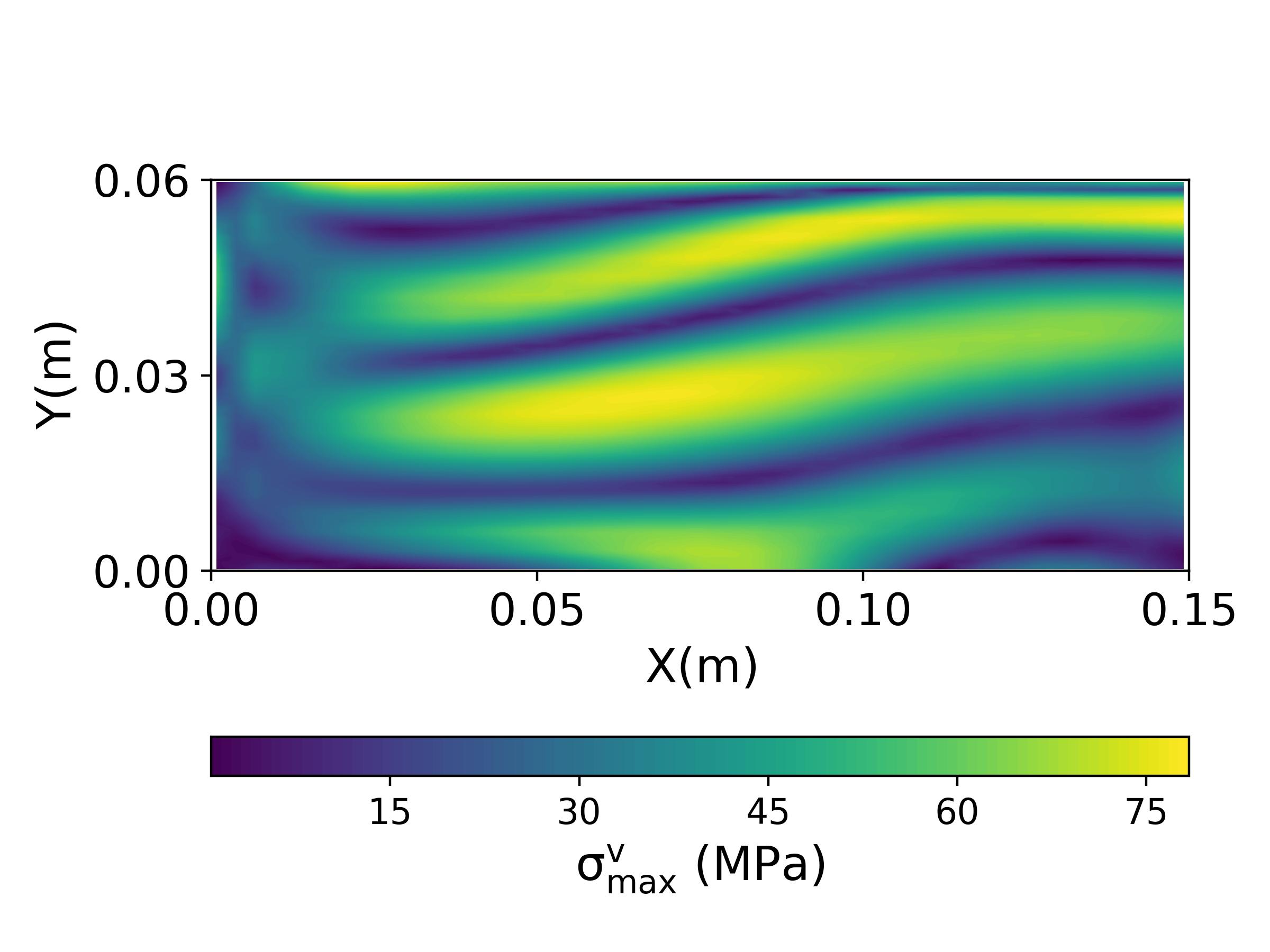}
        \caption{}
        \label{stress_1_2a}
    \end{subfigure}
    \begin{subfigure}[b]{0.33\textwidth}
        \centering
        \includegraphics[width=\textwidth]{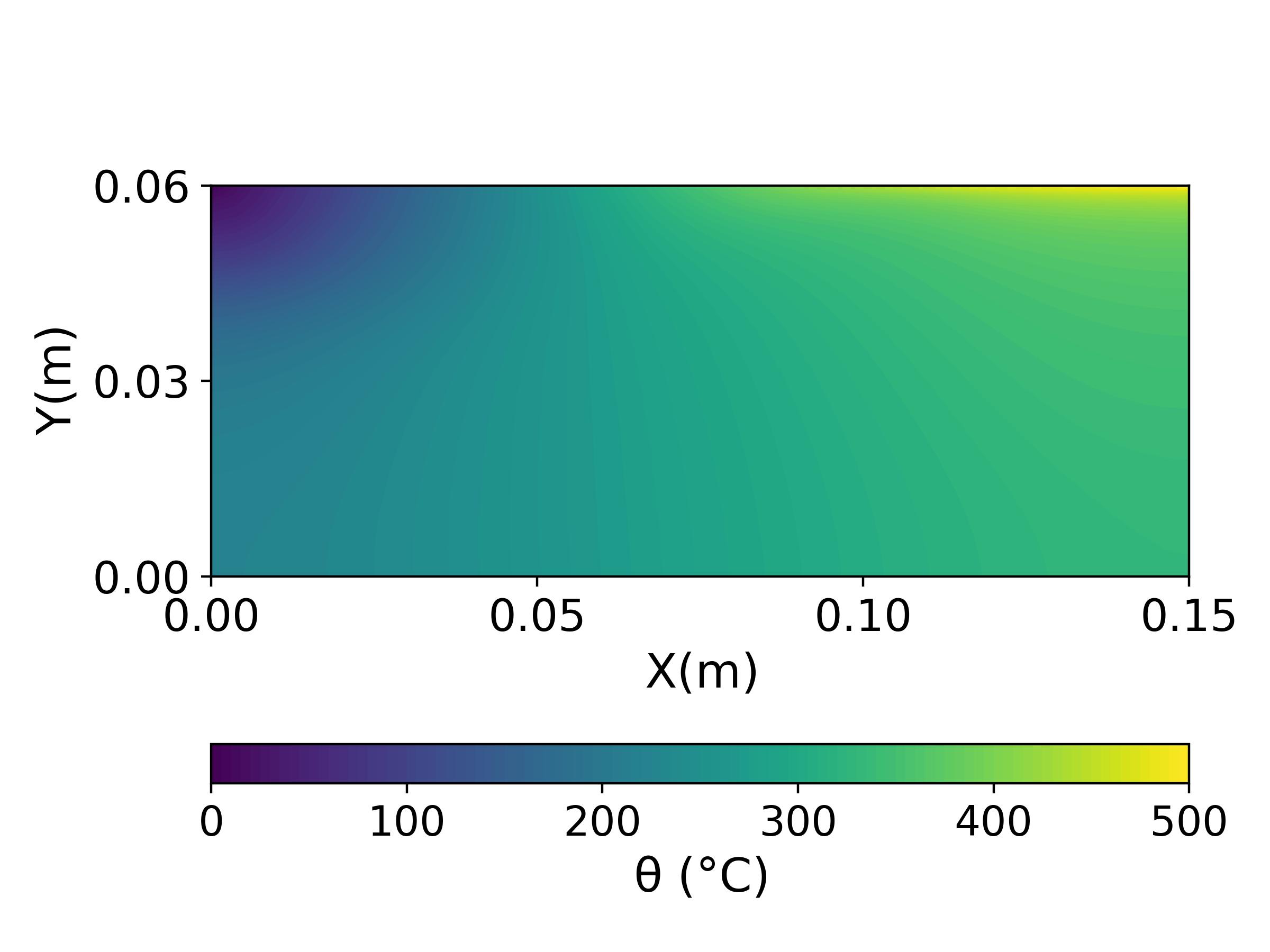}
        \caption{}
        \label{temp_1_2a}
    \end{subfigure}
    \caption{Optimized FGM plate under the constraint that the $V_{c}$ remains greater than 0.95 in regions where the $\theta$ exceeds 400\textdegree{C}: (a) contour plot of the ceramic volume fraction distribution, (b) corresponding von Mises stress distribution, and (c) temperature distribution.}
    \label{}
\end{figure*}

We run this problem with two different choices of the threshold temperature \(\theta^{*}\). In the first instance, the optimization was performed with a threshold temperature $\theta^{*} = \SI{400}{\degreeCelsius}$ and a prescribed minimum ceramic volume fraction $V_{c}^{*} = 0.95$. The resulting optimized volume fraction distribution is shown in Fig. \ref{vof_1_2a}, corresponding to a $\sigma_{max}^{v}$ of 78 MPa. The associated stress distribution is presented in Fig. \ref{stress_1_2a}. From Figs.~\ref{vof_1_2a} and \ref{temp_1_2a}, high ceramic content can be observed in regions where the temperature exceeds \SI{400}{\degreeCelsius}, aligning with the imposed constraint. The increase in the ceramic content can also be observed with respect to the unconstrained optimization by comparing Fig.~\ref{vof_1_2a} with Fig.~\ref {vof_1_1a}. 

\begin{figure*}[ht]
    \centering
    \begin{subfigure}[b]{0.33\textwidth}
        \centering
        \includegraphics[width=\textwidth]{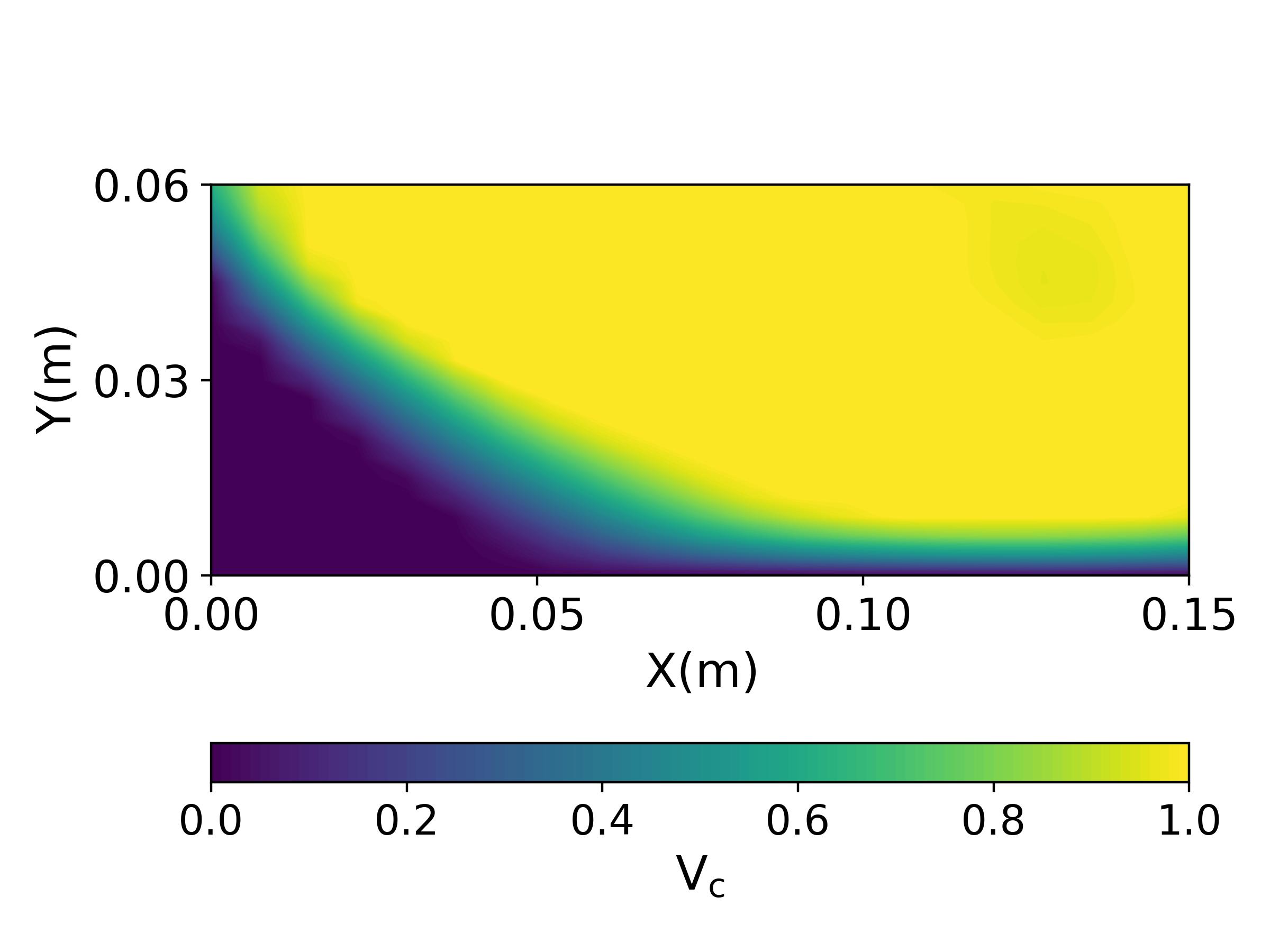}
        \caption{}
        \label{vof_1_2b}
    \end{subfigure}
    \begin{subfigure}[b]{0.33\textwidth}
        \centering
        \includegraphics[width=\textwidth]{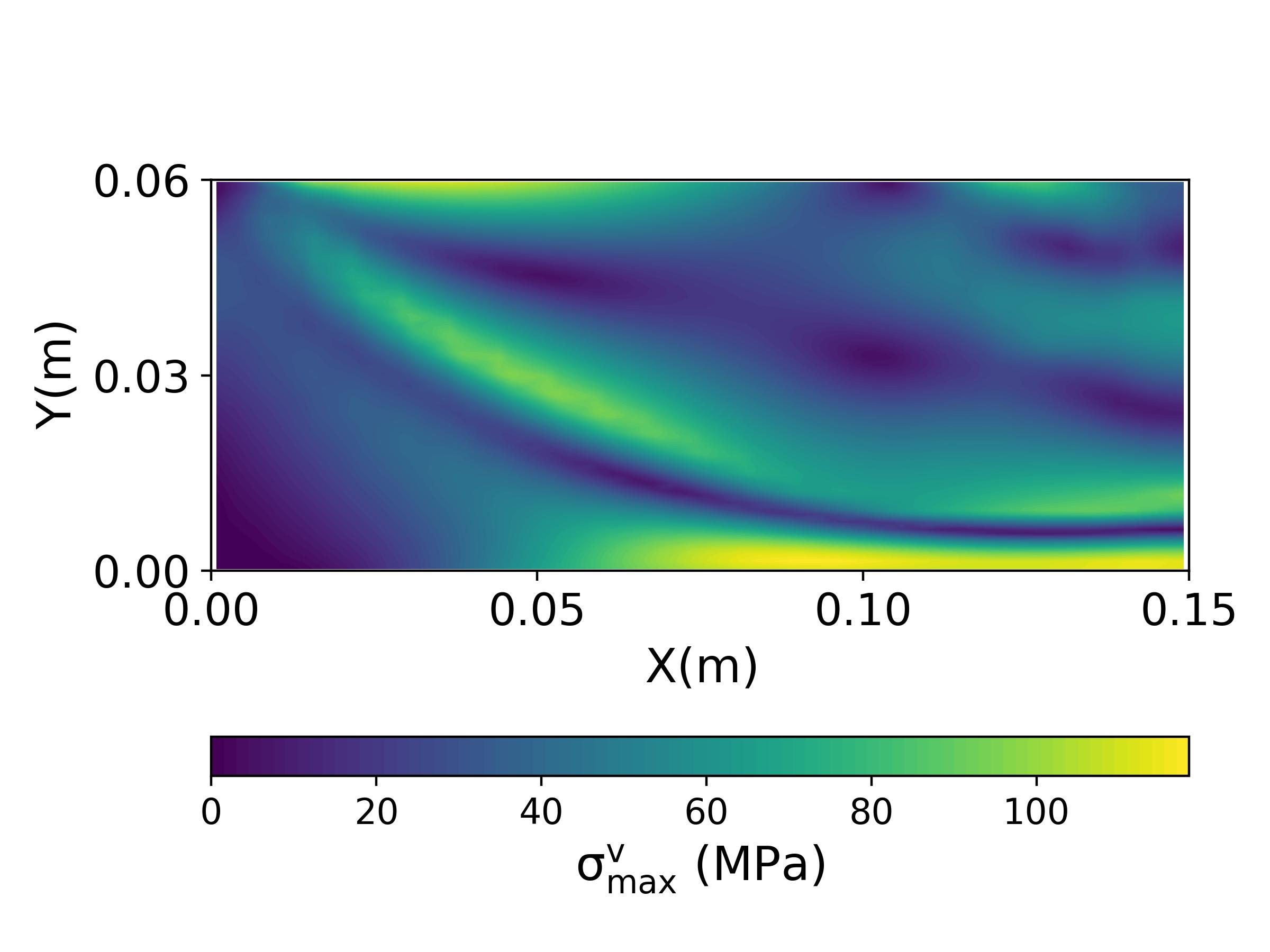}
        \caption{}
        \label{stress_1_2b}
    \end{subfigure}
    \begin{subfigure}[b]{0.33
    \textwidth}
        \centering
        \includegraphics[width=\textwidth]{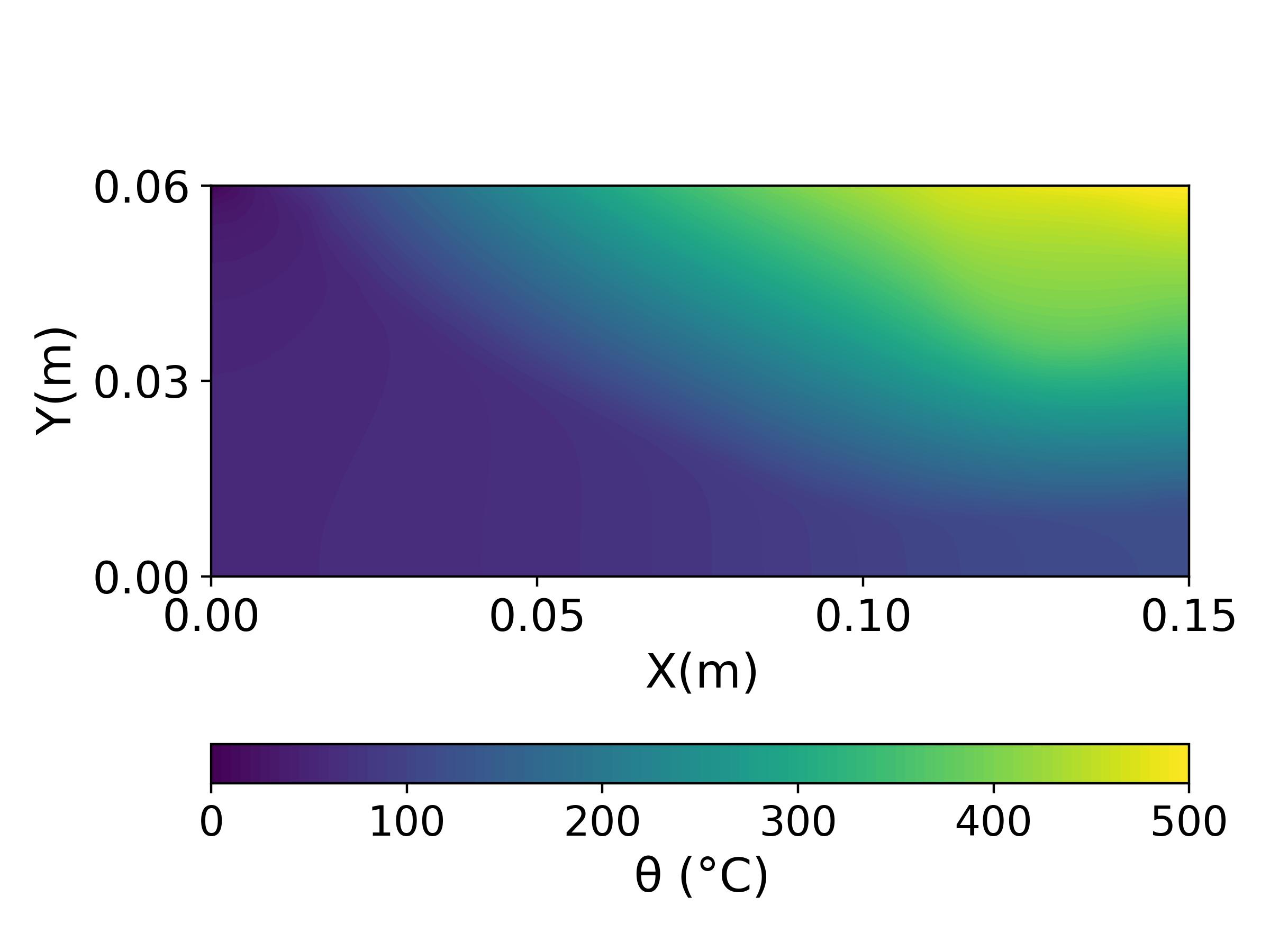}
        \caption{}
        \label{temp_1_2b}
    \end{subfigure}
    \caption{Optimized FGM plate under the constraint that the $V_{c}$ remains greater than 0.95 in regions where the $\theta$ exceeds 300\textdegree{C}: (a) contour plot of the ceramic volume fraction distribution, (b) corresponding von Mises stress distribution, and (c) temperature distribution.}
    \label{}
\end{figure*}

In the second case, a modified constraint has been applied: if the temperature within the FGM plate exceeds \SI{300}{\degreeCelsius}, the corresponding ceramic volume fraction must be greater than 0.95. The resulting optimized FGM profile, along with the von Mises stress field and temperature field, is illustrated in Figs. \ref{stress_1_2b} and \ref{temp_1_2b}, respectively. The volume fraction distribution for this case is shown in Fig. \ref{vof_1_2b}, where the $\sigma_{max}^{v}$ is 118 MPa.

The comparison of the optimum profile in Fig. \ref{vof_1_2b}, with Fig. \ref{vof_1_2a}, shows the increased ceramic content due to tighter thermal constraint. Further, as expected, enforcing stricter constraints leads to an increased value of the optimum stress. The increase in $\sigma_{max}^{v}$ from 78 MPa to 118 MPa highlights the trade-off between mechanical performance and thermal protection. 

\subsection{Problem 2: Square FGM plate with a circular hole}
\begin{figure}[htbp]
  \centering
  \includegraphics[width=0.42\textwidth]{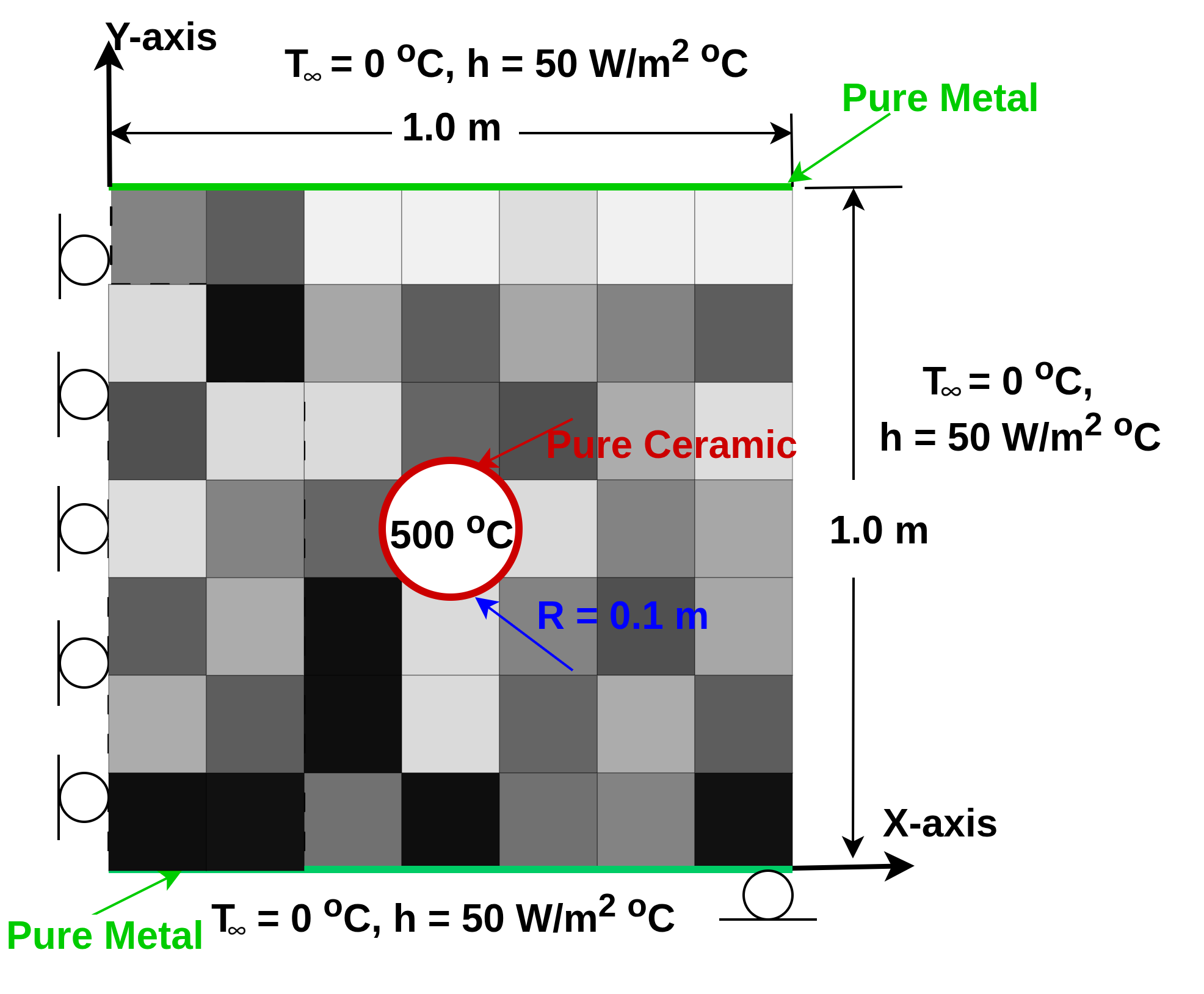}
  \caption{Problem 2: Schematic of square FGM plate with a circular hole, subjected to high temperature loading at the periphery of the hole.}
  \label{prb2_diagram}
\end{figure}

In this problem,  we increase the complexity of the problem by introducing a circular hole at the center of the plate. We consider one half of a simply supported FGM plate, featuring a circular hole, as illustrated in Fig.~\ref{prb2_diagram} under plane stress conditions. Temperature of \SI{500}{\degreeCelsius} is applied along the periphery of the circular hole. The left edge of the square plate is subjected to an adiabatic boundary condition ($q = 0$), and the displacement along the X-axis is zero ($u_1 = 0$). The top, bottom, and right edges of the plate are subjected to convective heat transfer and traction-free boundary conditions. The convective heat transfer coefficient is $h = \SI{50}{\watt\per\metre\squared\per\degreeCelsius}$, and $\theta_{\infty}$ is maintained at \SI{0}{\degreeCelsius}. The parameters used for GPR are $l$ = 0.3 and $\sigma$ = 1, and the GA parameters are given in Table \ref{tab:P2_GA_Para}.

The design constraint on the volume fraction requires a pure ceramic region at the periphery of the hole, while the top and bottom edges are constrained to a pure metal region. For the FEA, we utilized a total of 480 nine-noded quadrilateral elements. The termination criteria for the GA remain unchanged from the preceding example. The objective of the optimization is to find the profile with the minimum value of the \(\sigma_{max}^{v}\) as stated below:
\begin{equation}
\textbf{Minimize}: \quad \sigma_{max}^{v}(V^i_{c}), \quad i=1,2,..,n. 
\end{equation}

\begin{table}
    \renewcommand{\arraystretch}{1.3}
    \centering
        \caption{Parameters of the genetic algorithm.}
    \begin{tabular}{p{4.0cm}p{4.0cm}}
        \hline
        \textbf{Parameter} & \textbf{Value} \\
        \hline
        Population size & 200 \\
        Tournament size & 4 \\
        Crossover parameter & $\eta = 1.5\left[1+\tfrac{1}{2}\left(1-e^{-g/100}\right)\right]$ \\
        Mutation parameters & $l = 0.3$, $\sigma = 1$ \\
        Mutation probability & 0.3 \\
        \hline
    \end{tabular}
    \label{tab:P2_GA_Para}
\end{table}

\begin{figure}
    \centering
    
    \begin{subfigure}{\textwidth}
        \centering
        \begin{subfigure}{0.9\textwidth}
\includegraphics[trim=0 30 0 30,clip,width=\linewidth]{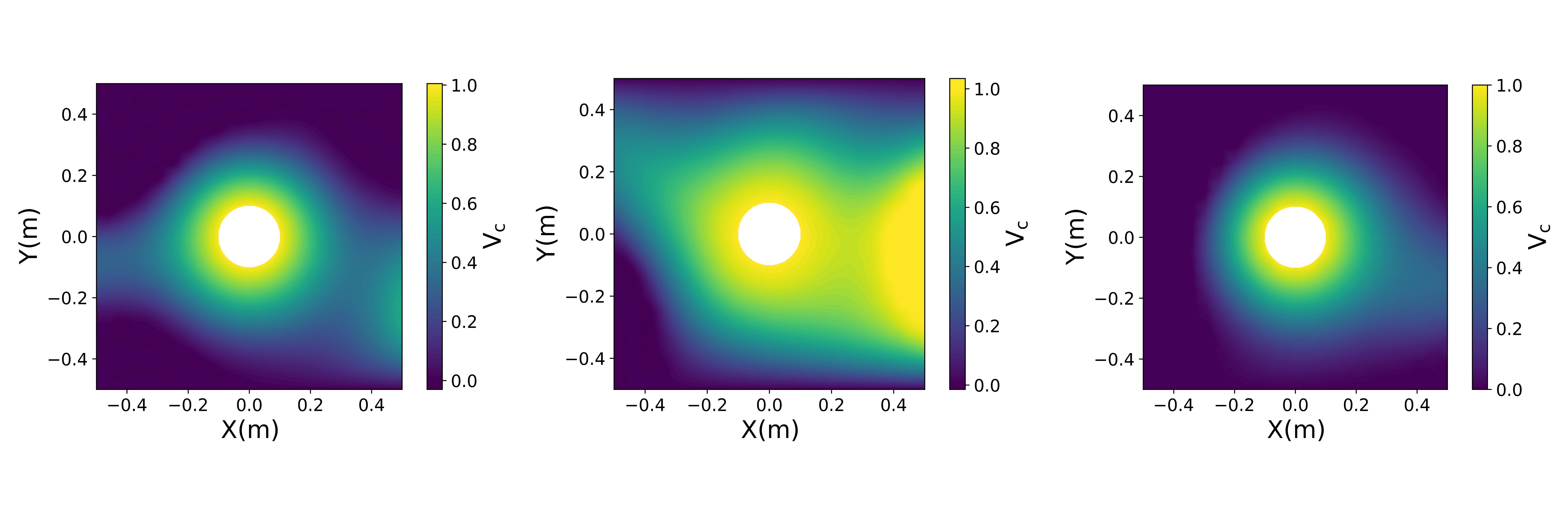}
        \end{subfigure}
        \caption{}
        \label{circle_prof}
    \end{subfigure}
    
    \vspace{0.1em} 
    
    \begin{subfigure}{\textwidth}
        \centering
        \begin{subfigure}{0.9\textwidth}
            \includegraphics[trim=0 30 0 30,clip,width=\linewidth]{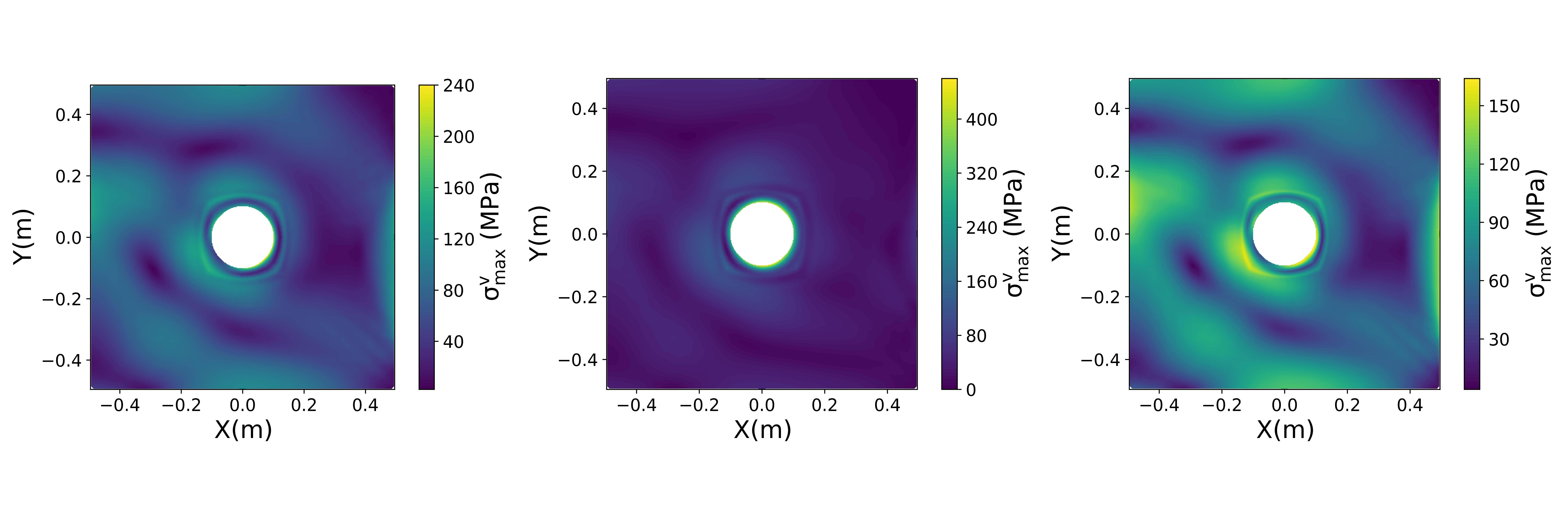}
        \end{subfigure}
        \caption{}
        \label{circle_stress}
    \end{subfigure}
    
    \caption{Sample FGM profiles for the square plate having a circular hole at the center generated through GPR-based profile generation algorithm: (a) contour plot of the ceramic volume fraction distribution and (b) corresponding von Mises stress distribution. }
    \label{prob2_random}
\end{figure}

\begin{figure}[h!]
    \centering
    \begin{subfigure}[b]{0.3\textwidth}
        \centering
        \includegraphics[width=\textwidth]{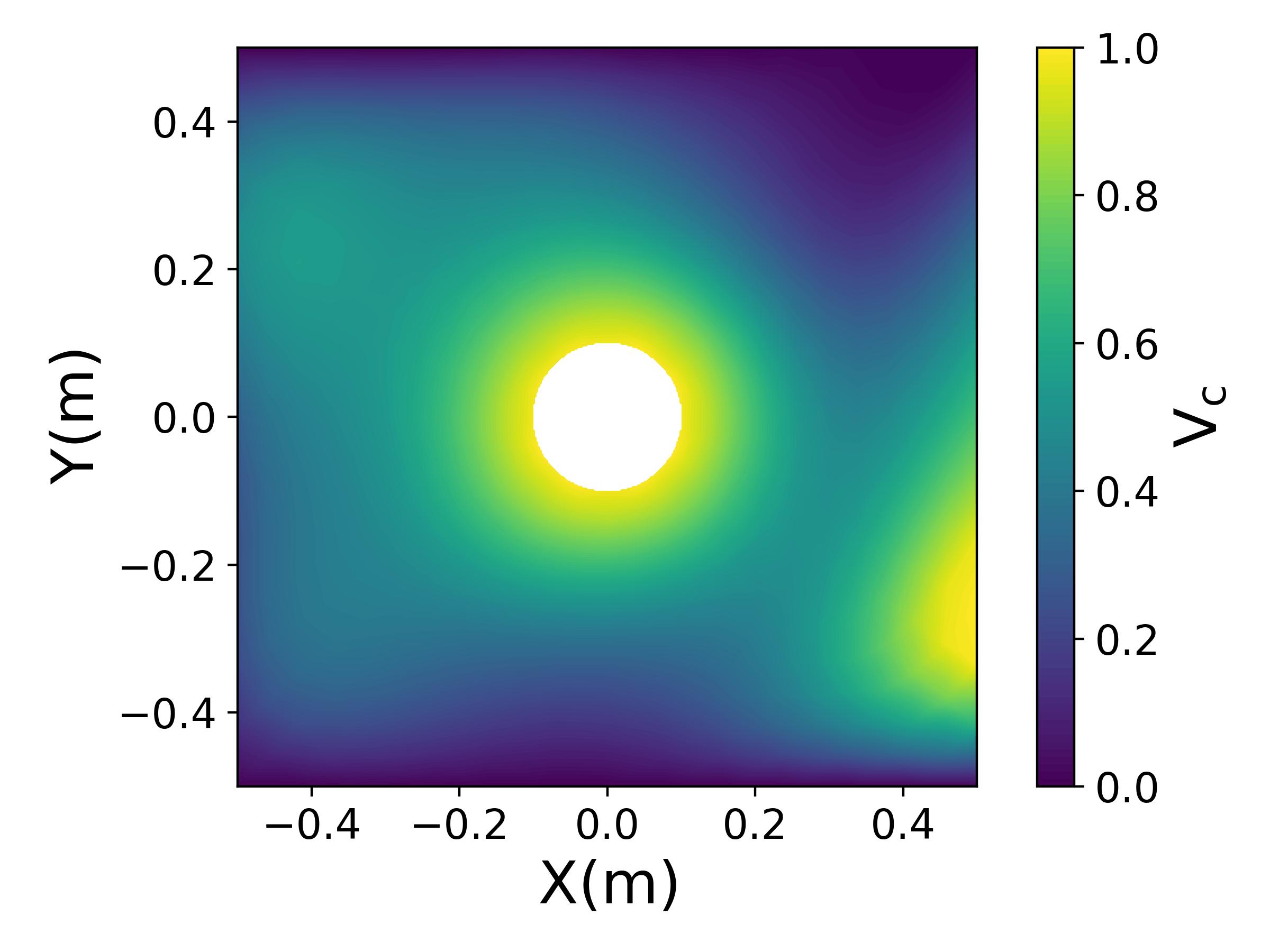}
        \caption{}
        \label{vof_2}
    \end{subfigure}
    \begin{subfigure}[b]{0.3\textwidth}
        \centering
        \includegraphics[width=\textwidth]{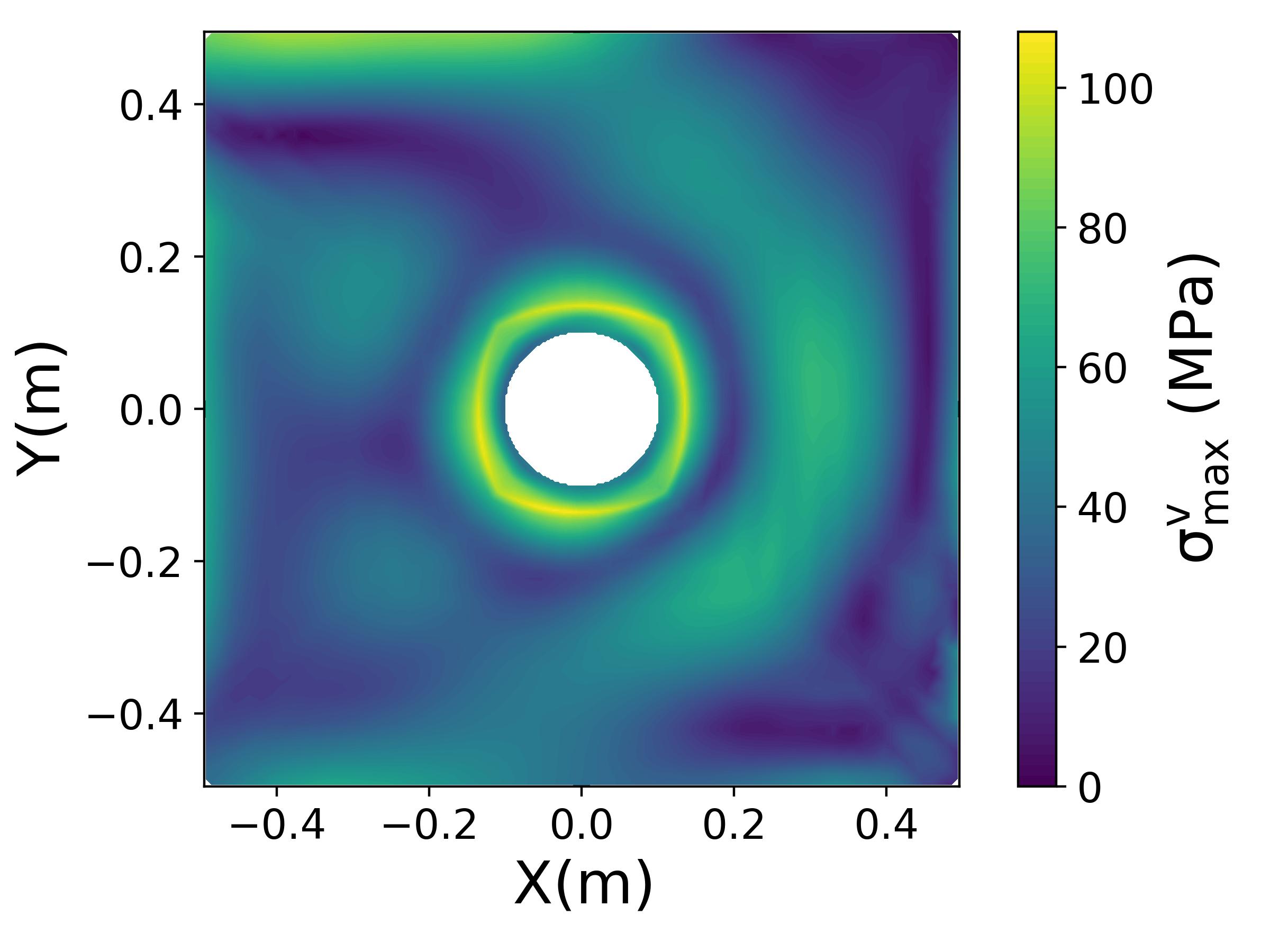}
        \caption{}
        \label{stress_2}
    \end{subfigure}
    \begin{subfigure}[b]{0.3\textwidth}
        \centering
        \includegraphics[width=\textwidth]{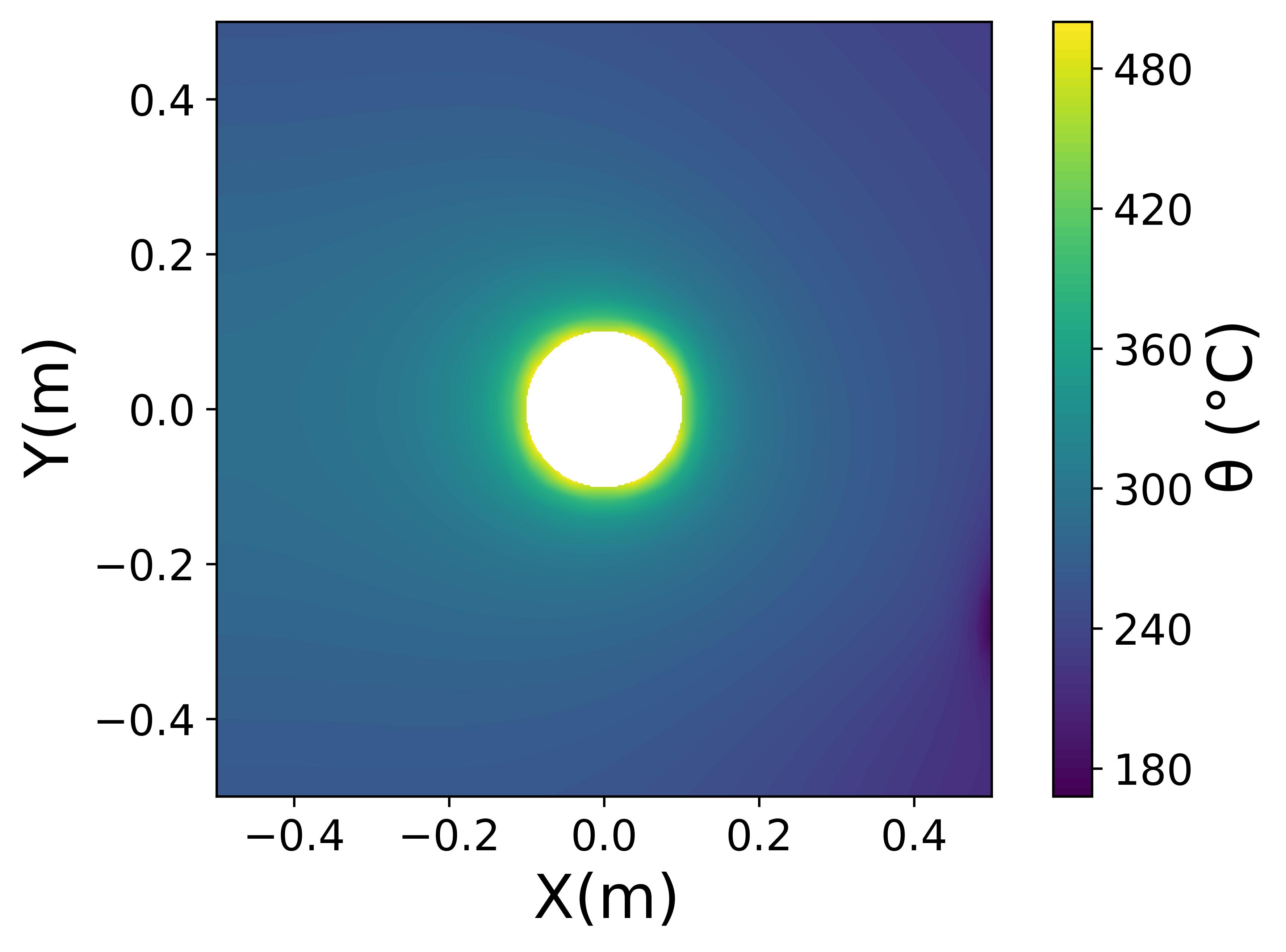}
        \caption{}
        \label{temp_2}
    \end{subfigure}
    \caption{Optimized FGM plate with circular hole at the center: (a) contour plot of the ceramic volume fraction distribution, (b) corresponding von Mises stress distribution, and (c) temperature distribution.}
    \label{}
\end{figure}

\begin{figure}[h!]
  \centering
  \includegraphics[width=0.37\textwidth]{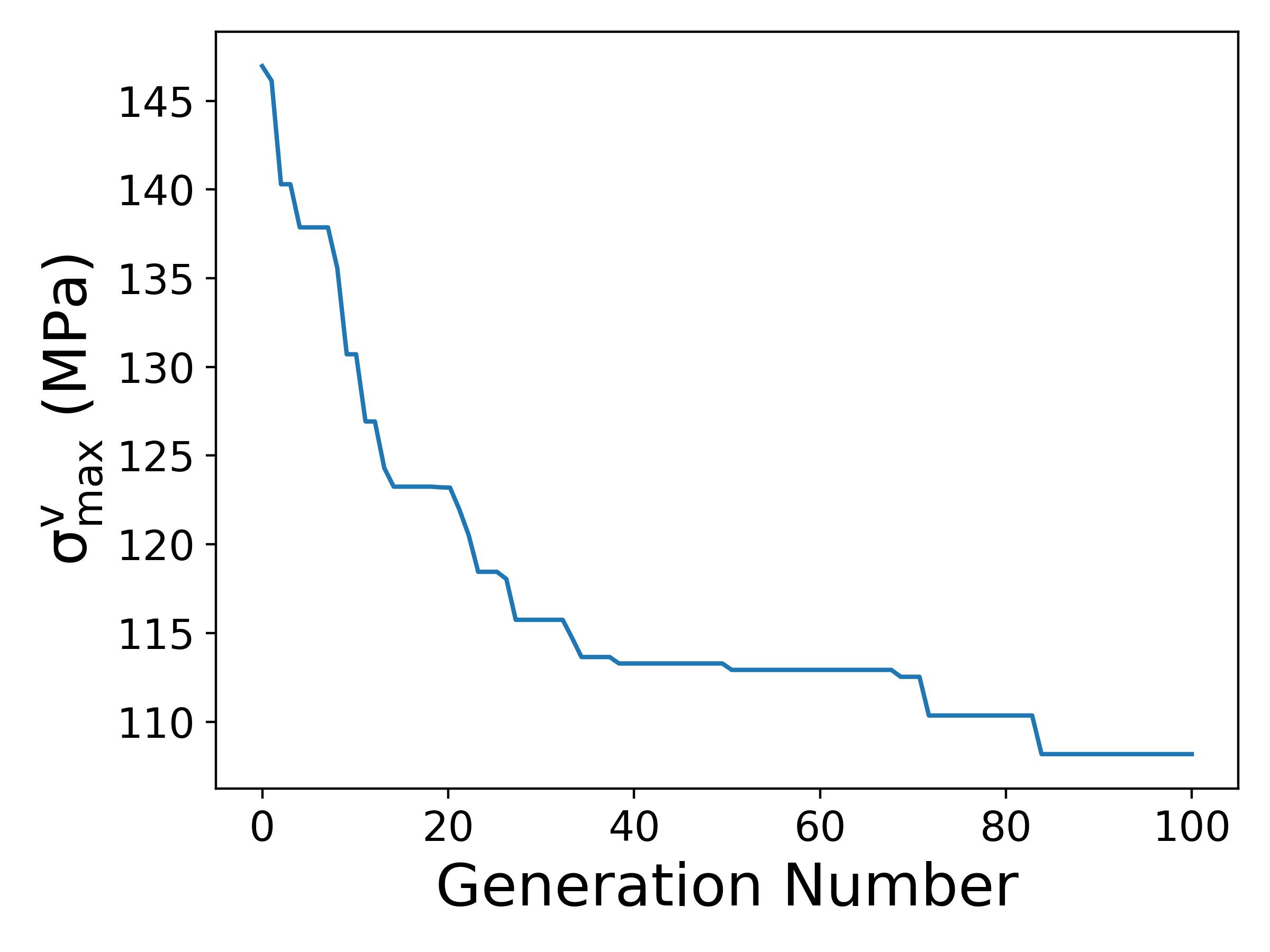}
  \caption{Evolution of the best fitness value over the generations during the optimization of a square FGM plate having a circular hole at the center.}
  \label{prb2_convergence}
\end{figure}

A couple of the randomly generated profiles from the GPR and the corresponding von Mises stress distribution are shown in Fig. \ref{circle_prof} and Fig. \ref{circle_stress}, respectively. These profiles do satisfy the constraint of the specified volume fraction on the boundary and are smooth in nature. However, we can observe the presence of high stresses near the hole for both of these random profiles. The optimum profile obtained from the optimization framework, along with its corresponding stress and temperature distributions, is shown in Fig. \ref{vof_2}, \ref{stress_2}, and \ref{temp_2}, respectively. The optimal value of $\sigma_{max}^{v}$ obtained in the optimized FGM profile is 108 MPa. As can be seen from the figures (Fig.~\ref{circle_stress} and Fig.~\ref{stress_2}), compared to the random profile, the optimum profile has a considerable reduction in the stress intensity near the circular hole region. Lastly, the evolution of $\sigma_{max}^{v}$ of the best individual for each generation is shown in the Fig. \ref{prb2_convergence}.

\subsection{Problem 3: Half elliptical FGM plate with two circular holes}
\begin{figure*}
  \centering
  \includegraphics[width=0.52\textwidth]{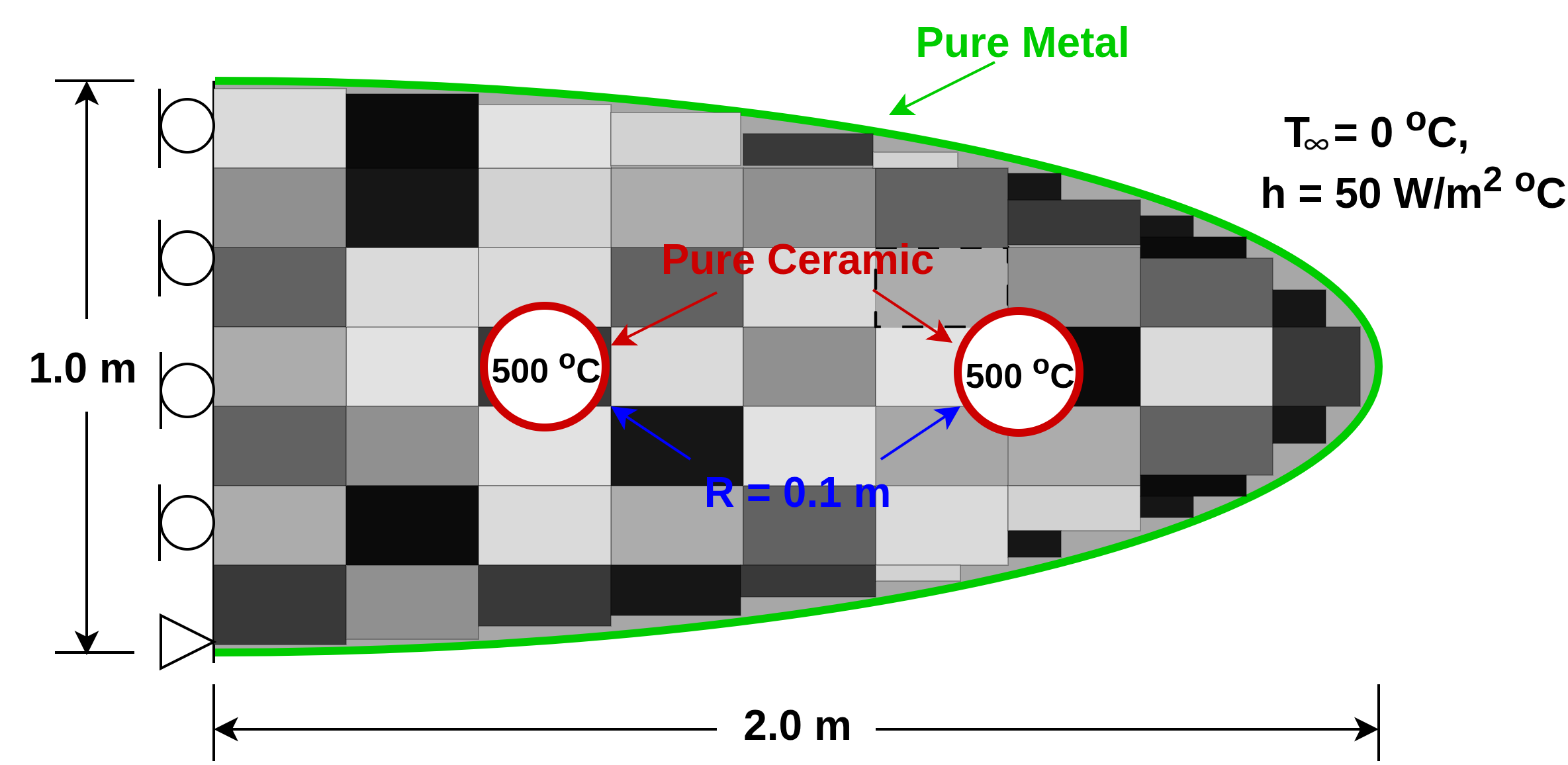}
  \caption{Problem 3: Schematic of half elliptical $ \mathrm{Al} / \mathrm{ZrO_2} $ FGM plate with two circular holes, subjected to the high temperature loading at the periphery of the holes.}
  \label{prb3_diagram}
\end{figure*}

In this numerical example, we consider the geometry of the half elliptical plate with two holes as illustrated in the Fig. \ref{prb3_diagram}. This geometry is significantly more complex than the geometries studied in the previous examples. The design constraints on the volume fraction are as follows: the periphery of both circular holes is constrained to be pure ceramic ($V_{c} = 1$), while the curved edge of the plate is constrained to be pure metal ($V_{c} = 0$). This thermo-elastic problem is considered as plane stress, and the setup for this example is defined as follows: the left (straight) edge is subjected to an adiabatic ($q = 0$) and roller support ($u_{1} = 0$) boundary conditions. Additionally, the bottom left corner is constrained with a fixed boundary condition. The curved edge of the plate is subjected to the convective heat transfer condition, with the convection parameter being the same as in the previous problem, while the peripheries of the two circular holes are maintained at a constant high temperature of \SI{500}{\degreeCelsius}. The ambient temperature is assumed to be \SI{0}{\degreeCelsius}. The finite element mesh employed for the analysis consists of 640 nine-noded quadrilateral elements.

In this problem, we consider two optimization cases: (1) minimization of maximum $\sigma_{max}^{v}$, (2) minimization of $V_{c}$ with specified $\sigma_{max}^{v}$ as a constraint. The termination criteria used are as follows: (1) a minimum number of 100 generations, and (2) the improvement in the objective value over the last 10 generations must be less than 0.01 MPa for the case of $\sigma_{\text{max}}^{v}$ minimization; whereas for volume fraction minimization the change in the average volume fraction over the last 10 generations must be less than 0.001. We have taken other  GA Parameters, the same as the previous problem.

\subsubsection{Case 1}
In this case, our objective is to find the optimal material distribution for minimum  $\sigma_{max}^{v}$. The optimization problem can be stated as follows:

\begin{equation}
\textbf{Minimize}: \quad \sigma_{max}^{v}(V^i_{c}), \quad i=1,2,..,n. 
\end{equation}

\begin{figure}
    \centering
    
    \begin{subfigure}{\textwidth}
        \centering
        \begin{subfigure}{0.9\textwidth}
            \includegraphics[trim=0 30 0 30,clip,width=\linewidth]{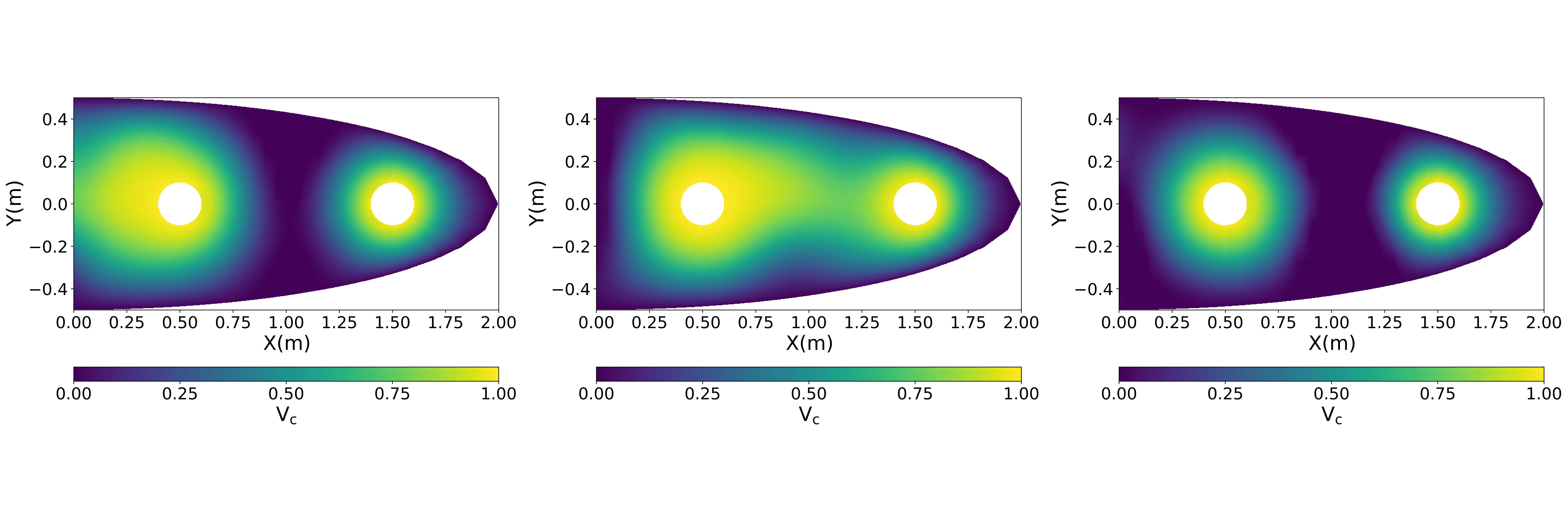}
        \end{subfigure}
        \caption{}
        \label{half_ellipse_prof}
    \end{subfigure}
    
    \vspace{0.1em} 
    
    \begin{subfigure}{\textwidth}
        \centering
        \begin{subfigure}{0.9\textwidth}
            \includegraphics[trim=0 30 0 30,clip,width=\linewidth]{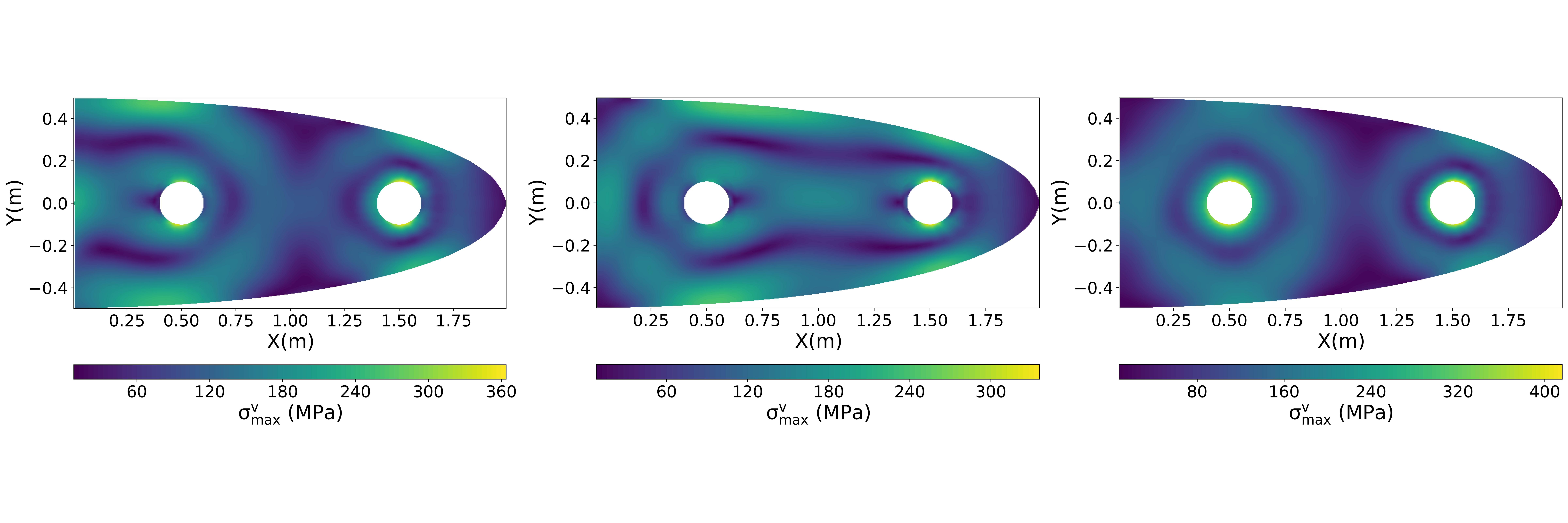}
        \end{subfigure}
        \caption{}
        \label{half_ellipse_stress}
    \end{subfigure}

    \caption{Sample FGM profiles for the half-elliptical plate having two circular holes generated through GPR-based profile generation algorithm (a) contour plot of the ceramic volume fraction distribution and (b) corresponding von Mises stress distribution. }
     \label{half_ellipse_random_stress}
\end{figure}

The design space for this case has been generated with GPR \(l= 0.3\) and \(\sigma=1.0\). We show two of the random profiles generated by GPR  Fig. \ref{half_ellipse_prof}. As can be seen from the figure, the boundary conditions are strictly satisfied, with a pure ceramic phase ($V_{c} = 1$) along the periphery of the internal holes, and a pure metal phase ($V_{c} = 0$) along the curved edge of the plate. These profiles also exhibit a smooth and continuous transition from ceramic to metal throughout the entire domain. The von Mises stress distribution corresponding to these profiles is shown in ~\ref{half_ellipse_stress}. In these random profiles, the high intensity of the stress near the circular region can be observed. 

\begin{figure}[h!]
    \centering
    \begin{subfigure}[b]{0.33\textwidth}
        \centering
        \includegraphics[width=\textwidth]{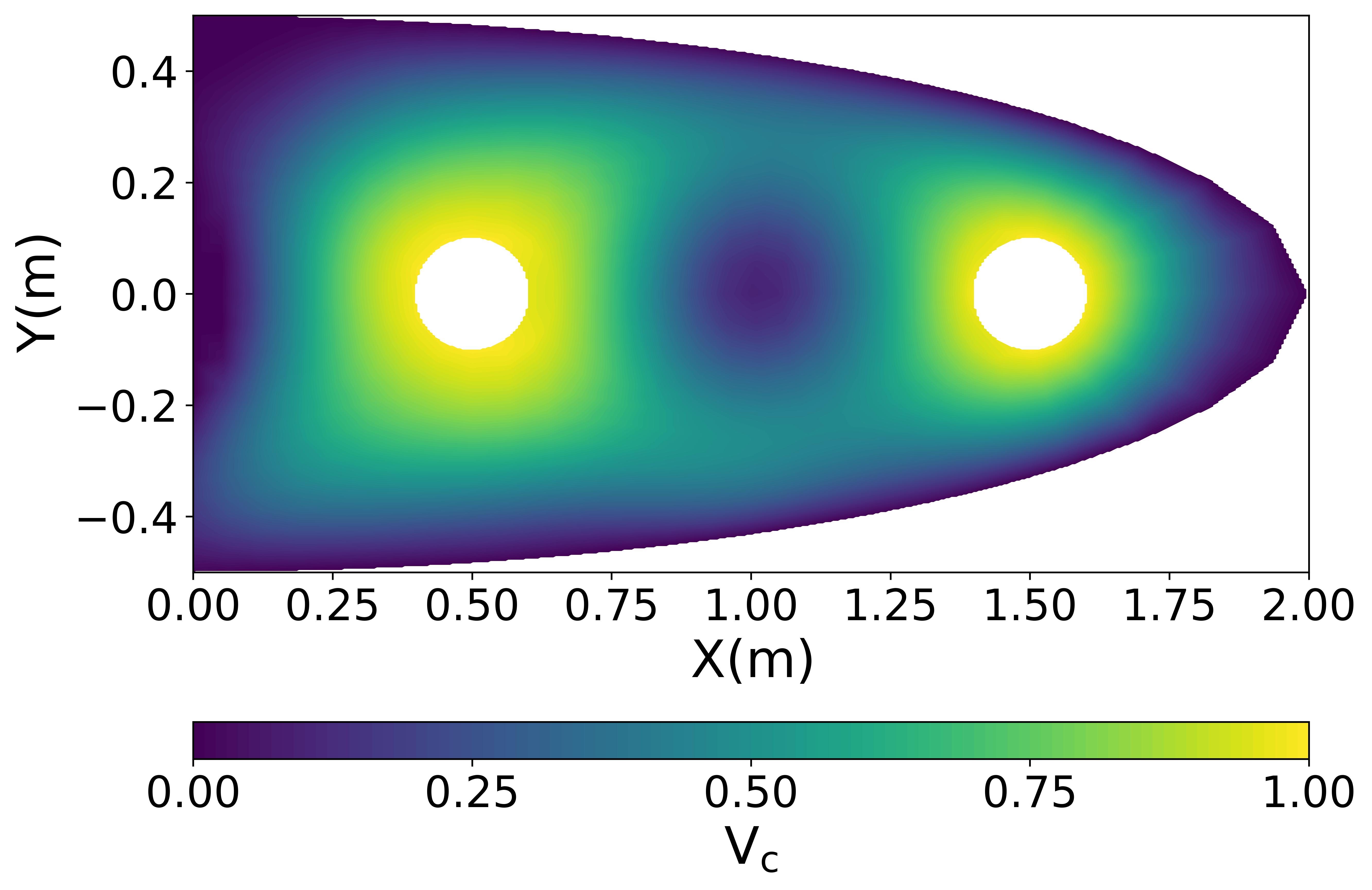}
        \caption{}
        \label{vof_3a}
    \end{subfigure}
    \begin{subfigure}[b]{0.33\textwidth}
        \centering
        \includegraphics[width=\textwidth]{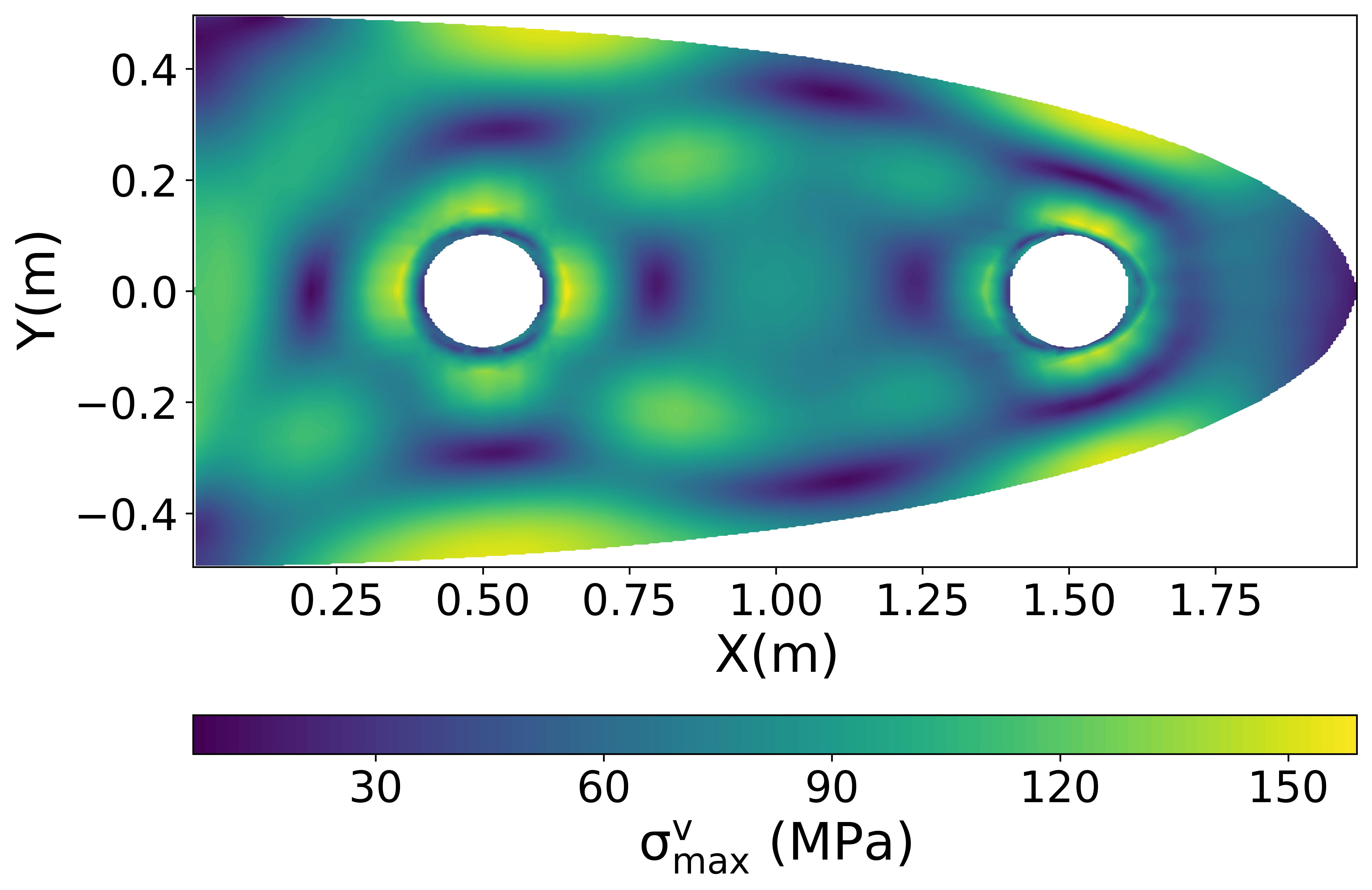}
        \caption{}
        \label{stress_3a}
    \end{subfigure}
    \begin{subfigure}[b]{0.33\textwidth}
        \centering
        \includegraphics[width=\textwidth]{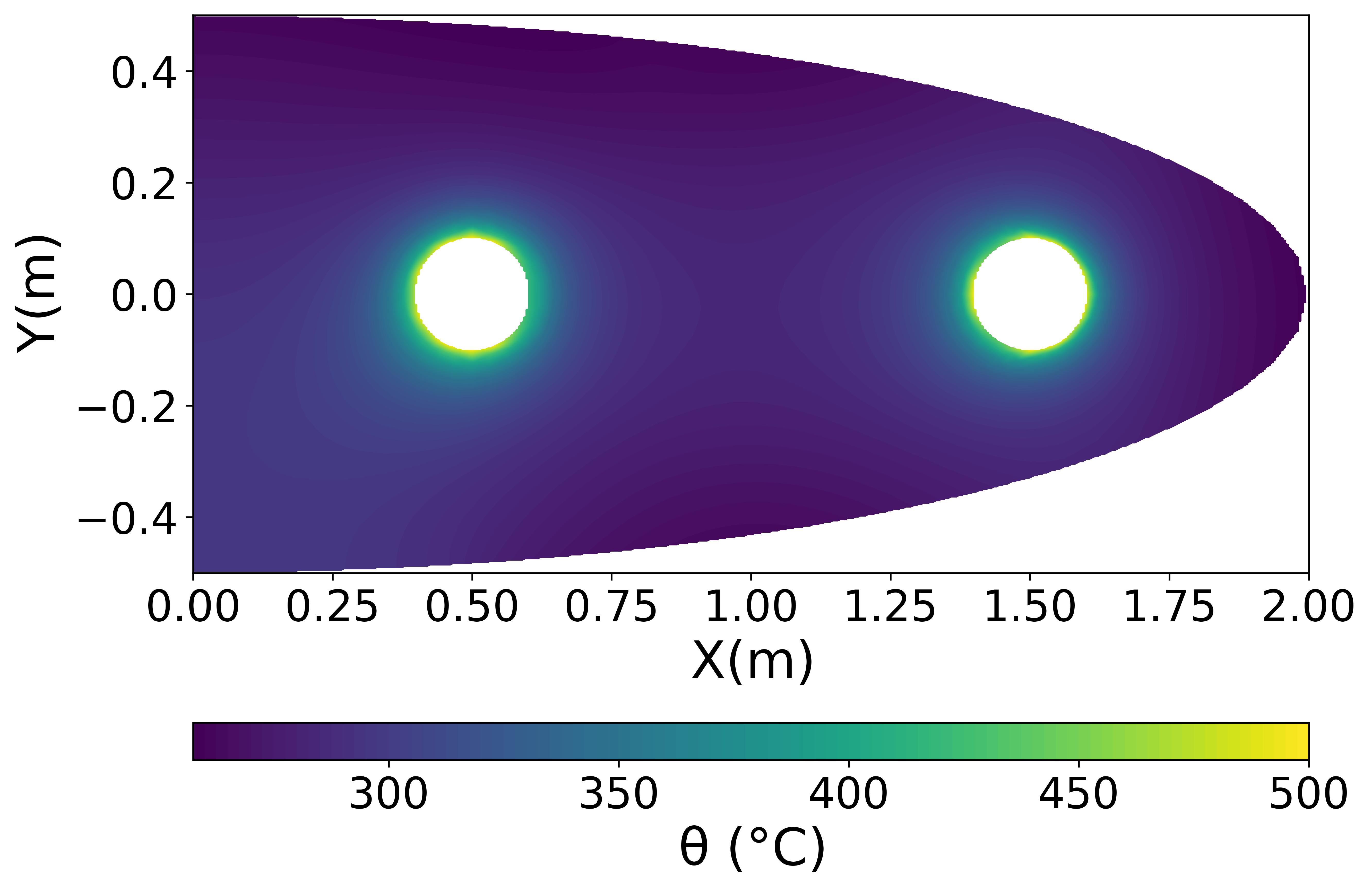}
        \caption{}
        \label{temp_3a}
    \end{subfigure}
    \caption{Optimized half ellipse FGM plate with two circular holes: (a) contour plot of the ceramic volume fraction distribution, (b) corresponding von Mises stress distribution, and (c) temperature distribution.}
    \label{}
\end{figure}

The obtained optimized profile is illustrated in Fig.~\ref{vof_3a}. The corresponding  Von Mises stress distribution is shown in the Fig. \ref{stress_3a}, while the temperature distribution in the optimized profile is given in the Fig. \ref{temp_3a}. The maximum value of Von Mises stress is 158 MPa. As can be seen from the stress distribution, the stress distribution in the case of the optimum profile is significantly more uniform compared to the random profile cases. Thus, the GPR-based algorithm proves to be an effective and robust method for generating FGM distributions within arbitrarily shaped domains, while also accommodating volume fraction constraints.

\subsubsection{Case 2}
In this case, we are performing a constrained optimization problem with the objective of minimizing the ceramic volume fraction under the constraint of the $\sigma_{max}^{v}$. The optimization problem is stated as follows:

\begin{equation}
\begin{aligned}
&\textbf{Minimize:} \quad && V^i_{c}, \quad i = 1,2,\dots,n, \\
&\textbf{Subject to:} \quad &&  \sigma_{max}^{v}(V^i_{c}) \le \sigma^{*}.
\end{aligned}
\end{equation}

\begin{figure}[h!]
    \centering
    \begin{subfigure}[b]{0.33\textwidth}
        \centering
        \includegraphics[width=\textwidth]{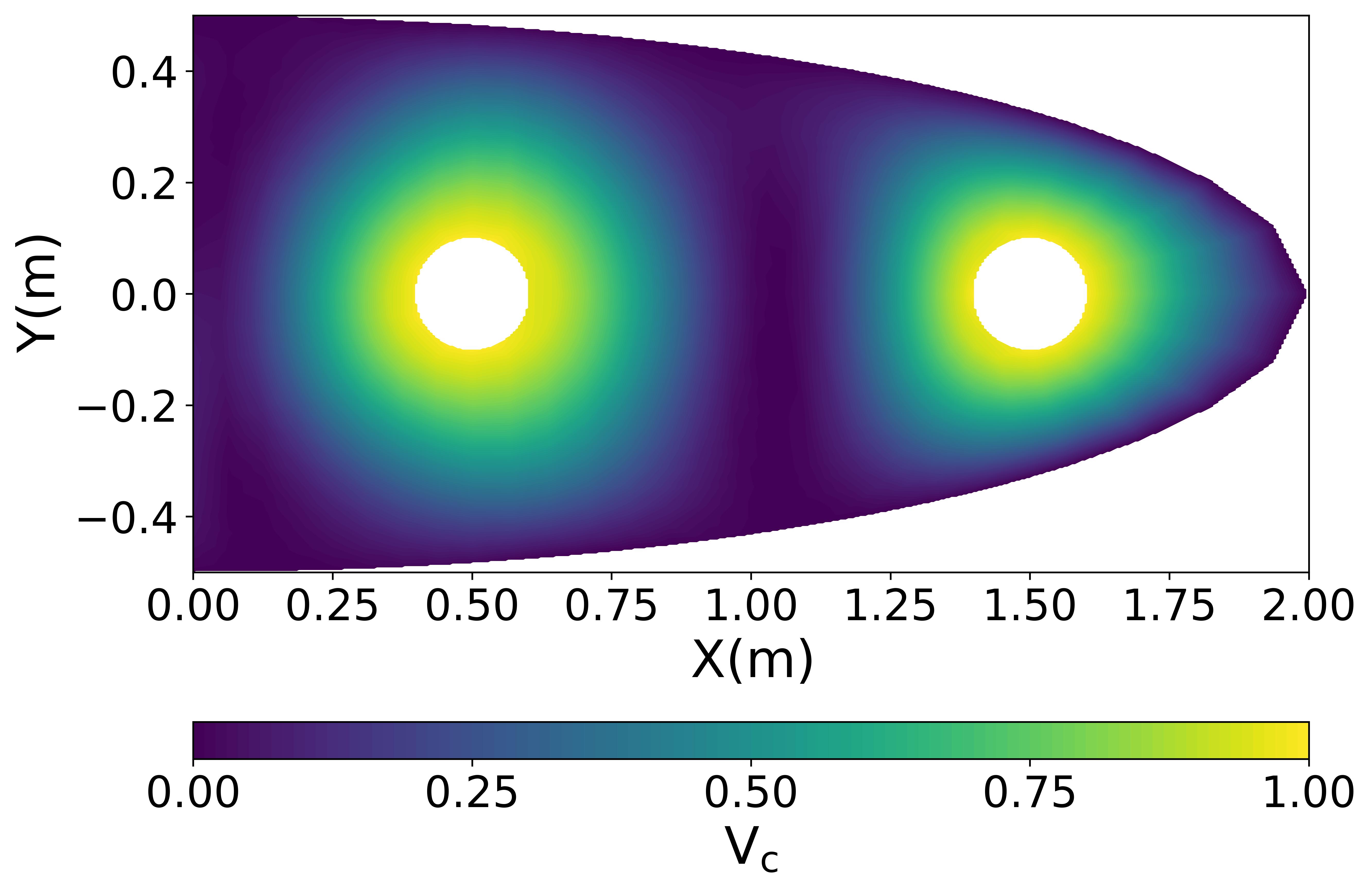}
        \caption{}
        \label{vof_3b}
    \end{subfigure}
    \begin{subfigure}[b]{0.33\textwidth}
        \centering
        \includegraphics[width=\textwidth]{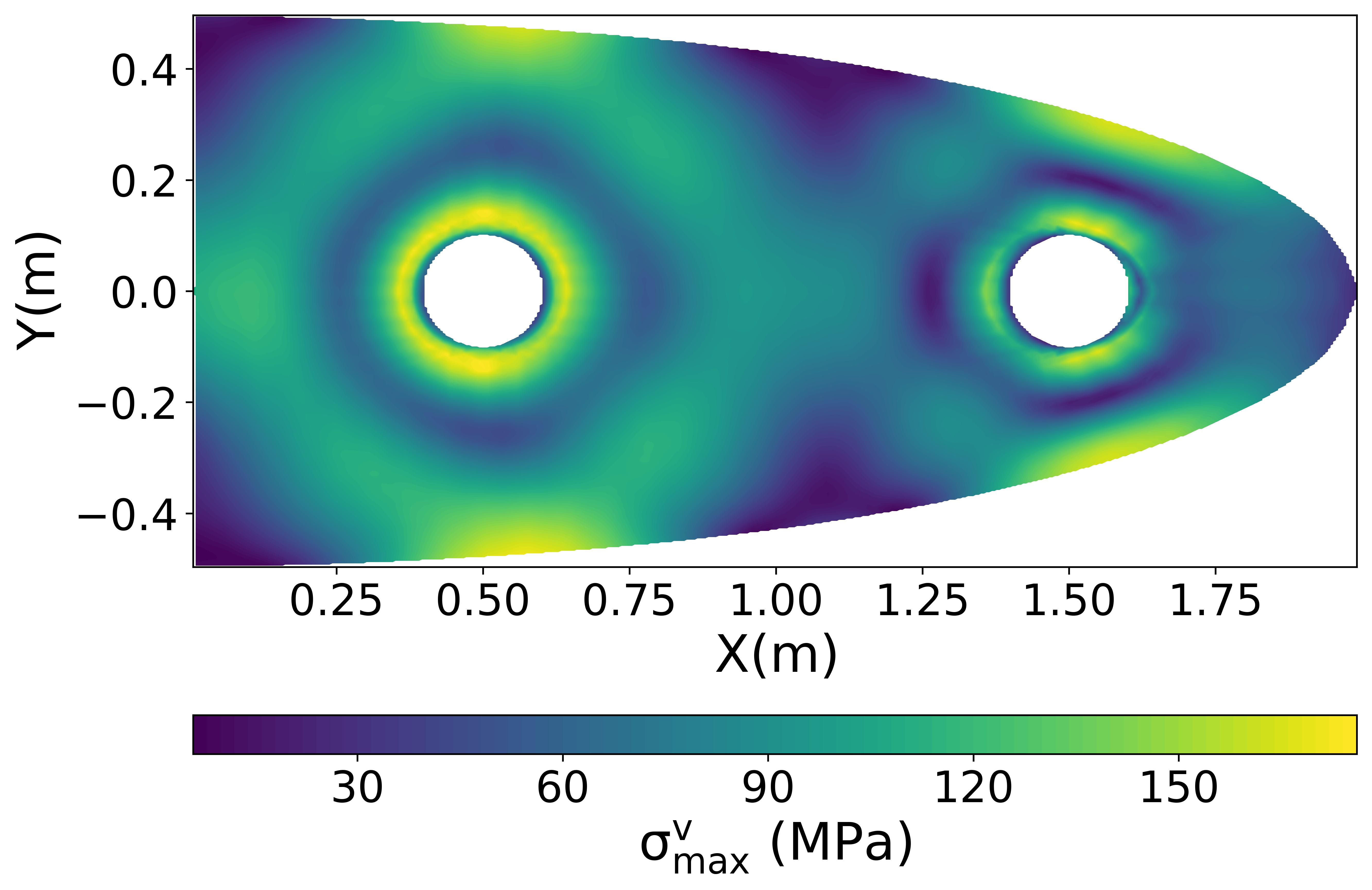}
        \caption{}
        \label{stress_3b}
    \end{subfigure}
    \begin{subfigure}[b]{0.33\textwidth}
        \centering
        \includegraphics[width=\textwidth]{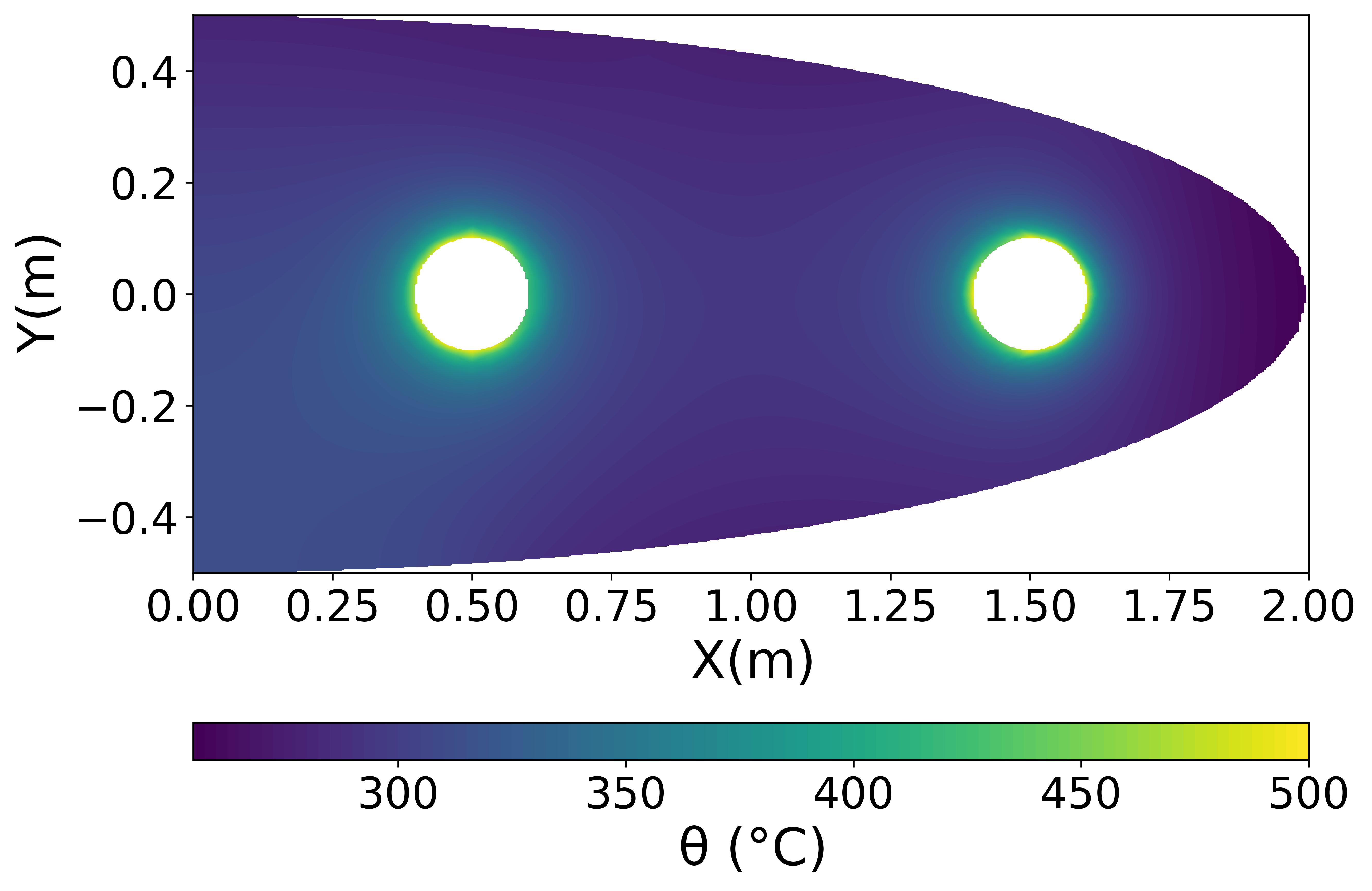}
        \caption{}
        \label{temp_3b}
    \end{subfigure}
    \caption{Optimized half ellipse FGM plate with two circular holes under the $\sigma^{v}_{max}$ constraint condition: (a) contour plot of the ceramic volume fraction distribution, (b) corresponding von Mises stress distribution, and (c) temperature distribution.}
    \label{}
\end{figure}

For this case, a threshold value of $\sigma^{*}$ = 175 MPa is taken. We observed the stress values lie in the range 180 MPa- 600 MPa, for 10,000 randomly generated samples. Thus, the selected value is slightly lower compared to the minimum of this observed range. The optimum FGM profile for this stress constraint case has been shown in Fig. \ref{vof_3b}. The optimized profile has a ceramic content of 0.33353. Further, the temperature and von Mises stress distribution are given in the Figs. \ref{stress_3b} and \ref{temp_3b}. The satisfaction of the stress constraint can be observed from the distribution.

Designing FGM profiles for irregular geometries with specific constraints remains a challenging task. Consequently, the existing literature offers limited studies that explore optimization beyond regular or standard geometries. However, the proposed GPR-based profile generation algorithm significantly simplifies this process, enabling the design of arbitrary geometries while effectively incorporating various design constraints.

\section {Conclusion}
In the manuscript, we have proposed a profile generation algorithm based on the Gaussian process regression. The proposed algorithm has been shown to generate smooth FGM profiles while adhering to the specified volume fraction values at boundaries/part of the boundaries for various complex-shaped domains. The smoothness of the generated profiles and the size of the underlying design space have been shown to be controlled by a length scale parameter. Further, the effective integration of the proposed profile generation algorithm with the genetic algorithm (GA) has been carried out. The rich design space from the profile generation algorithm, in conjunction with the robust GA framework, results in optimum profiles towards the given objective while meeting the specified constraints. The same has been demonstrated through various thermoelastic non-constrained/constrained optimization problems. While the current study focuses on thermomechanical problems, the proposed scheme is generic and can be explored in other FGM applications as well.

\bibliographystyle{elsarticle-num}
\bibliography{refrence}

\end{document}